\pdfoutput=1
\documentclass{article}

\usepackage[preprint]{neurips_2025}

\usepackage{enumitem}

\usepackage{times}
\usepackage{hyperref}

\usepackage[T1]{fontenc}

\usepackage[utf8]{inputenc}
\usepackage{microtype}
\usepackage{inconsolata}
\usepackage{graphicx}
\usepackage{amsmath}
\usepackage{colortbl}

\usepackage{wasysym}
\usepackage{amssymb}
\usepackage{multirow}
\usepackage{booktabs}
\usepackage{tikz}
\usepackage[most]{tcolorbox}
\usepackage{booktabs}
\usepackage{array}
\usepackage{longtable}
\usepackage[utf8]{inputenc}
\usepackage[T1]{fontenc}

\usepackage{circledsteps}
\usepackage{fontawesome5}
\usepackage{fmtcount}

\usepackage{subfigure}

\usepackage{hyperref} 
\usepackage{url} 
\usepackage{booktabs} 
\usepackage{amsfonts} 
\usepackage{nicefrac} 
\usepackage{microtype} 
\usepackage{xcolor} 
\usepackage{listings}

\usepackage{wrapfig}
\usepackage{xcolor} 
\definecolor{myblue}{rgb}{0.4, 0.7, 1.0} 
\definecolor{mintgreen}{RGB}{152,255,152} 
\definecolor{lightmint}{RGB}{200,255,200} 

\usepackage[utf8]{inputenc}
\DeclareUnicodeCharacter{2212}{\ensuremath{-}}

\newcommand{\MMMR}{\textbf{MMLU-Reason}}
\definecolor{lightgray}{gray}{0.95}
\definecolor{lightred}{RGB}{255,232,232}
\definecolor{darkred}{RGB}{220,20,60}

\definecolor{darkgreen}{RGB}{54,179,54} 
\definecolor{lightgreen}{RGB}{224,249,224}
\newcommand{\colorred}[1]{\colorbox{lightred}{#1}}
\newcommand{\colorgreen}[1]{\colorbox{lightgreen}{#1}}

\title{MMLU-Reason: Benchmarking Multi-Task Multi-modal Language Understanding and Reasoning}

\author{
{\bfseries Guiyao Tie\textsuperscript{1}}\quad
{\bfseries Xueyang Zhou\textsuperscript{1}}\quad
{\bfseries Tianhe Gu\textsuperscript{1}}\quad
{\bfseries Ruihang Zhang\textsuperscript{1}}\quad
{\bfseries Chaoran Hu\textsuperscript{1}}\quad
{\bfseries Sizhe Zhang\textsuperscript{1}}\quad\\
{\bfseries Mengqu Sun\textsuperscript{2}}\quad
{\bfseries Yan Zhang\textsuperscript{1}}\quad
{\bfseries Pan Zhou\textsuperscript{1}}\quad
{\bfseries Lichao Sun\textsuperscript{2}} 
\\
{\bfseries \textsuperscript{1}Huazhong University of Science and Technology}\quad
{\bfseries \textsuperscript{2}Lehigh University}\\
\texttt{\{tgy,d202480819,u202211961,u202211917,u202314532,U202312332\}@hust.edu.cn}\\
\texttt{mes225@lehigh.edu,\{u202312543,panzhou\}@hust.edu.cn,lis221@lehigh.edu}
}

\begin{document}
  \maketitle

  \begin{abstract}
Recent advances in Multi-Modal Large Language Models (MLLMs) have enabled unified processing of language, vision, and structured inputs, opening the door to complex tasks such as logical deduction, spatial reasoning, and scientific analysis. Despite their promise, the reasoning capabilities of MLLMs—particularly those augmented with intermediate thinking traces (MLLMs-T)—remain poorly understood and lack standardized evaluation benchmarks. Existing work focuses primarily on perception or final answer correctness, offering limited insight into how models reason or fail across modalities. To address this gap, we introduce the \MMMR, a new benchmark designed to rigorously evaluate multi-modal reasoning with explicit thinking. The \MMMR~comprises 1) a high-difficulty dataset of 1,083 questions spanning six diverse reasoning types with symbolic depth and multi-hop demands and 2) a modular Reasoning Trace Evaluation Pipeline (RTEP) for assessing reasoning quality beyond accuracy through metrics like relevance, consistency, and structured error annotations. Empirical results show that MLLMs-T overall outperform non-thinking counterparts, but even top models like Claude-3.7-Sonnet and Gemini-2.5 Pro suffer from reasoning pathologies such as inconsistency and overthinking. This benchmark reveals persistent gaps between accuracy and reasoning quality and provides an actionable evaluation pipeline for future model development. Overall, the \MMMR~offers a scalable foundation for evaluating, comparing, and improving the next generation of multi-modal reasoning systems. Project Page: \underline{\href{https://mmmr-benchmark.github.io/}{https://mmmr-benchmark.github.io/}}.
  \end{abstract}

\begin{figure}[h]
\centering
\includegraphics[width=1\textwidth]{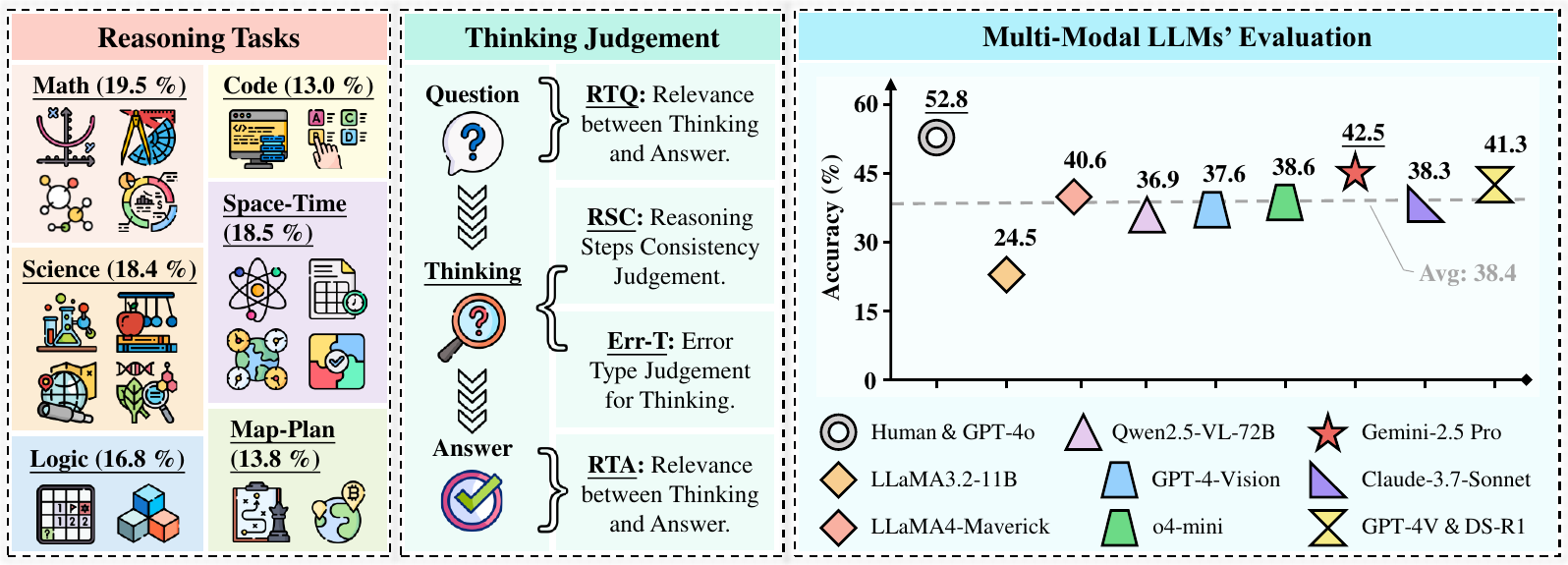}
\vspace{-15pt}
\caption{Overview of the \MMMR. Left: Six reasoning tasks. Middle: Thinking Judgements assessing alignment for relevance (RTQ, RTA), consistency (RSC), and error (Err-T). Right: Accuracy distribution shows a gap between humans and state-of-the-art MLLMs-T.}
\vspace{-10pt}
\label{fig:overview}
\end{figure}

\section{Introduction}
\label{sec:introduction}

The rapid advancement of Multi-Modal Large Language Models (MLLMs) has significantly enhanced unified reasoning capabilities across language, vision, and structured data modalities. Early MLLMs such as Qwen-VL~\cite{bai2023qwen}, LLaVA~\cite{liu2023llava}, and GPT-4 Vision~\cite{gpt4vision} have demonstrated impressive performance in perception-centric tasks, including visual question answering~\cite{antol2015vqa}, image captioning~\cite{sharma2018conceptual}, and grounded retrieval~\cite{radford2021learning}. However, their proficiency remains limited in tasks necessitating structured reasoning, symbolic abstraction, or sequential multi-step inference. To address this limitation, a new paradigm—MLLMs incorporating explicit intermediate reasoning (MLLMs-T)—has emerged. Representative models such as Gemini-2.5 Pro~\cite{gemini25pro} and Claude-3.7-Sonnet~\cite{claude37sonnet} leverage Chain-of-Thought-style reasoning, decomposing complex problems into interpretable intermediate steps, thereby emulating human-like structured problem-solving in domains such as logical deduction, scientific analysis, and code reasoning.

Despite significant progress, rigorously evaluating the reasoning capabilities of MLLMs-T remains challenging. Current benchmarks, including MMBench~\cite{liu2023mmbench}, MME-CoT~\cite{mmecot}, and MMMU~\cite{mmmu}, predominantly emphasize broad coverage of tasks and perceptual understanding, offering limited insights into the reasoning process itself. These benchmarks primarily focus on answer correctness without assessing the underlying reasoning's consistency, coherence, or cognitive alignment. Consequently, a critical research question arises: \textbf{\textit{To what extent do MLLMs-T reliably generate coherent, interpretable, and cognitively aligned reasoning traces in complex multi-modal tasks?}}

Addressing this research question demands a benchmark emphasizing reasoning depth rather than breadth, evaluating not only final predictions but also the intermediate reasoning processes explicitly. Such a benchmark requires 1) challenging multi-modal reasoning tasks explicitly designed to probe structured inference capabilities, and 2) robust criteria for systematically evaluating intermediate reasoning quality. Currently, no benchmark adequately fulfills these requirements~\cite{liu2023mmbench, mmecot, mmmu, lu2024mathvista}, leaving a significant gap in diagnosing and attributing reasoning failures at the trace level.

To bridge this gap, we introduce \MMMR, a comprehensive benchmark explicitly designed to evaluate the multi-modal reasoning capabilities of both MLLMs and MLLMs-T. As shown in Figure~\ref{fig:overview}, our benchmark comprises 1,083 rigorously curated high-difficulty tasks spanning six reasoning domains: logical reasoning~\cite{mmecot,xiao2024logicvista}, mathematical problem-solving~\cite{lu2024mathvista,emma}, spatio-temporal understanding~\cite{mmecot,xu2024mcbench}, code reasoning~\cite{li2024web2code,li2024mmcode}, map-based planning~\cite{liu2024multimodalselfinstruct,liu2024mapevalvisual}, and scientific analysis~\cite{mmecot,mmmu}. Each task integrates diverse modalities, including text, images, tables, and diagrams, carefully designed to require structured, symbolic, and abstract reasoning beyond mere perception.

Unlike prior benchmarks~\cite{liu2023mmbench, mmmu,lu2024mathvista}, \MMMR~introduces a structured \textit{Reasoning Trace Evaluation Pipeline (RTEP)}, capturing reasoning trace relevance, logical consistency, and frequent error types. This structured evaluation identifies key reasoning pitfalls such as overthinking, trace inconsistency, and logical errors. The right panel of Figure~\ref{fig:overview} reveals a pronounced performance gap between state-of-the-art MLLMs-T and human-level expert reasoning. Specifically, while the best-performing model, Gemini-2.5 Pro, achieves a test accuracy of 42.45\%, human experts assisted by GPT-4o reach 52.85\%. This 10.3\% margin underscores the persistent challenge in closing the reasoning gap, even with advanced architectures and explicit thinking mechanisms.

In summary, our contributions include:
\begin{itemize}
\item \textbf{A comprehensive benchmark for multi-modal reasoning.}~We introduce \MMMR, the first benchmark that systematically targets \textit{multi-modal reasoning} across six domains—Logic, Math, Code, Map, Science, and Space-Time. Unlike prior datasets, \MMMR~emphasizes hard question solutions and cross-modal alignment to increase reasoning complexity.

\item \textbf{The first evaluation pipeline for thinking of MLLMs-T.}~We propose the Reasoning Trace Evaluation Pipeline (RTEP), the first framework to incorporate intermediate \emph{thinking trace} analysis into multi-modal reasoning evaluation. RTEP assesses reasoning relevance, stepwise consistency, and alignment, which enables deeper diagnostic insight beyond accuracy.

\item \textbf{Insights into reasoning capabilities and failures.} Through extensive evaluation on \MMMR, we find that state-of-the-art MLLMs-T achieve high answer accuracy across tasks, yet frequently produce flawed reasoning traces—exhibiting logical inconsistency or overthinking. These findings expose a critical misalignment between surface-level correctness and reasoning fidelity, offering new evaluation directions for future multi-modal model architecture.

\end{itemize}

  \section{MMMR: Benchmarking Massive Multi-Modal Reasoning Tasks}
\label{sec:mmmr}

\begin{figure}[t]
\centering
\includegraphics[width=1\textwidth]{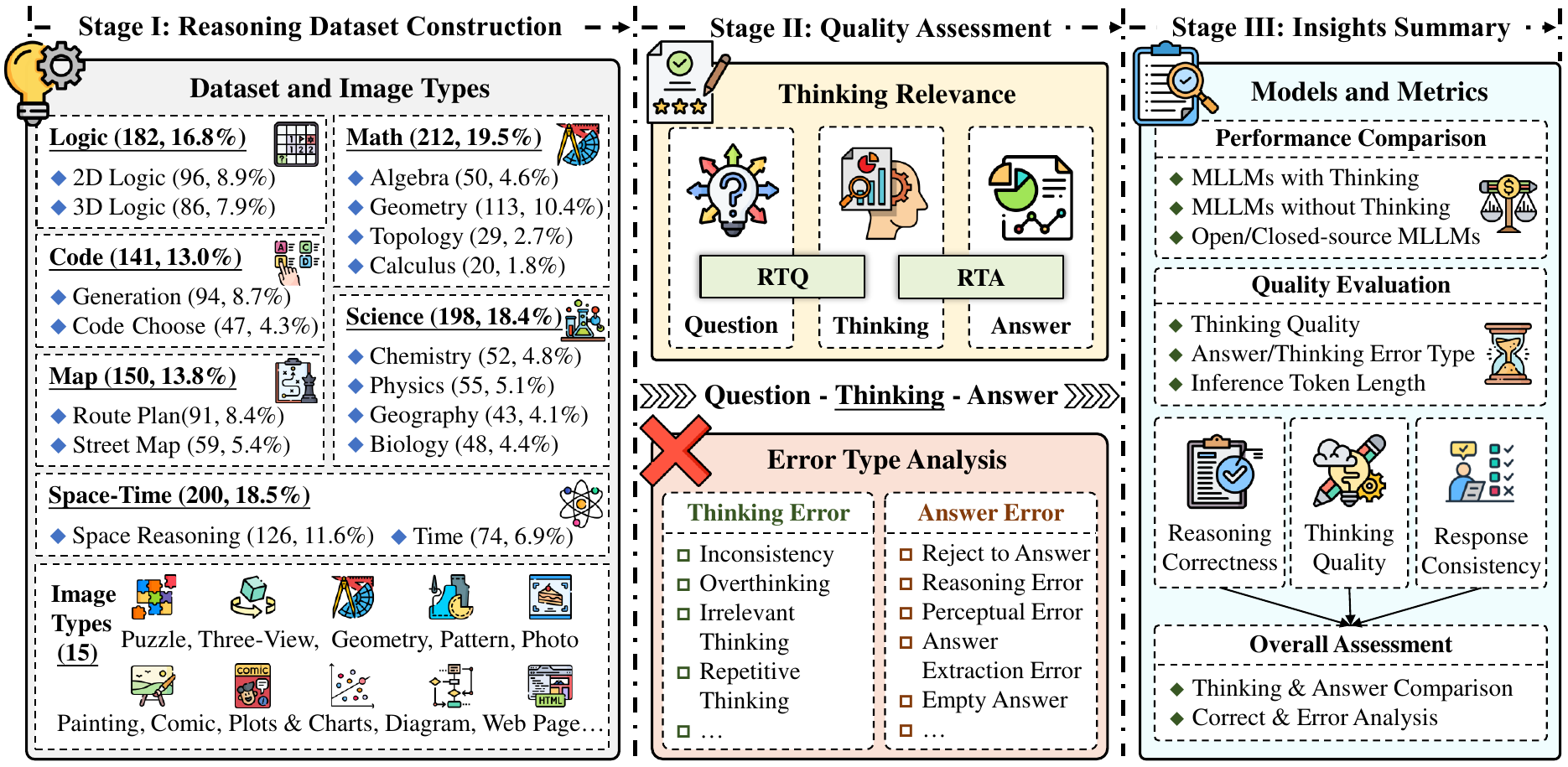}
\caption{Overview of the \MMMR~evaluation pipeline. Stage \uppercase\expandafter{\romannumeral1} involves the creation of a challenging multi-modal reasoning benchmark dataset. Stage \uppercase\expandafter{\romannumeral2} evaluates the quality and structural integrity of intermediate reasoning generated by MLLMs-T. Stage \uppercase\expandafter{\romannumeral3} synthesizes insights regarding reasoning strategies, effectiveness, and common failure patterns across different tasks and models.}
\label{fig:pipeline}
\end{figure}

The \MMMR~is a challenging benchmark dataset meticulously crafted to evaluate the reasoning capabilities of Multi-modal Large Language Models with intermediate Thinking (MLLMs-T). Unlike previous benchmarks~\cite{antol2015vqa,hudson2019gqa,johnson2017clevr,lu2022learn} which predominantly measure perception or general knowledge, \MMMR~emphasizes complex reasoning tasks requiring deep integration across diverse modalities, such as text, images, and structured data. Motivated by recent advancements in MLLMs-T, exemplified by Gemini-2.5 Pro~\cite{google2025gemini25}, which leverage intermediate reasoning processes to enhance performance, we propose a rigorous three-stage evaluation pipeline. This pipeline is specifically designed to evaluate multi-modal reasoning quality and, crucially, assess the effectiveness and robustness of intermediate thinking. As illustrated in Figure~\ref{fig:pipeline}, our evaluation pipeline comprises three core stages: (\uppercase\expandafter{\romannumeral1}) Reasoning Dataset Construction, (\uppercase\expandafter{\romannumeral2}) Thinking Quality Assessment, and (\uppercase\expandafter{\romannumeral3}) Reasoning Insights Synthesis.

\subsection{Stage \uppercase\expandafter{\romannumeral1}: Reasoning Dataset Construction}

In this initial stage, we construct the \MMMR~dataset to thoroughly evaluate MLLMs-T across a wide spectrum of reasoning scenarios. The  \MMMR~comprises 1,083 carefully curated multi-modal tasks, systematically categorized into six distinct reasoning types: Logic (16.8\%), Math (19.5\%), Space-Time (18.5\%), Code (13.0\%), Map (13.8\%), and Science (18.3\%). Each reasoning type further includes task-specific subcategories, such as deductive inference, algebraic calculation, temporal ordering, code generation, spatial planning, and hypothesis evaluation. The \MMMR~incorporates heterogeneous modalities including natural language texts, visual imagery, and structured data (e.g., Three-View diagrams, Plots \& Charts, and Web Pages). To facilitate reproducible and granular evaluation, the dataset is partitioned into a validation set (106 samples) and a test set (977 samples).

\subsection{Stage \uppercase\expandafter{\romannumeral2}: Thinking Quality Assessment}

This stage systematically evaluates the quality and structure of intermediate reasoning processes (i.e., \textit{thinking}) produced by MLLMs-T. We propose a reasoning trace evaluation pipeline (RTEP), including metrics: 1) RTQ, quantifying the relevance of \textit{thinking} with the posed question, and 2) RTA, assessing logical relevance between \textit{thinking} and the answer. Both metrics are normalized within the [0,1] interval and evaluated through standardized prompts designed for precise, unbiased assessment. Furthermore, we conduct an extensive error type analysis, categorizing reasoning failures into distinct types, including thinking errors and answer errors. This fine-grained analysis offers critical insights into the strengths and vulnerabilities inherent in the reasoning approaches adopted by MLLMs-T.

\subsection{Stage \uppercase\expandafter{\romannumeral3}: Reasoning Insights Synthesis}

The final stage synthesizes the observations from Stage \uppercase\expandafter{\romannumeral2} to generate holistic reasoning insights. We pursue three primary analytical objectives: 1) comparing and benchmarking the performance of MLLMs-T against standard MLLMs across different reasoning tasks, 2) profiling the quality of intermediate reasoning to identify consistent patterns of strength and weakness, and 3) investigating how prevalent error types (especially overthinking and redundant reasoning stages) impact the overall reliability of reasoning outcomes. By aggregating and analyzing detailed results across various tasks and model variants, this stage supports informed interpretation of reasoning behavior.

\subsection{Research Questions}

To systematically guide our evaluation and produce insightful analyses on the reasoning capabilities of MLLMs-T, we articulate the following research questions:

\begin{tcolorbox}[
  colback=red!3, 
  colframe=black, 
  arc=2mm, 
  boxrule=0.4mm,   
  width=\linewidth, 
  left=6pt, 
  right=6pt, 
  top=6pt, 
  bottom=6pt           
]
\textbf{[RQ1]} How do MLLMs-T perform in comparison to standard MLLMs concerning reasoning accuracy across the diverse and challenging multi-modal tasks presented in \MMMR?

\textbf{[RQ2]} How does the quality of intermediate thinking generated by MLLMs-T vary across different levels of task complexity and modality combinations?

\textbf{[RQ3]} Which reasoning error types are most frequently encountered by MLLMs-T within different task contexts of \MMMR, and how do these errors reflect underlying challenges in multi-modal integration?
\end{tcolorbox}
  \section{Experiment Settings}
\label{sec:setting}

\noindent\textbf{Multi-Modal Language Models without Thinking (MLLMs).}~~MLLMs solve multi-modal tasks by directly mapping perception inputs to answers, bypassing explicit reasoning steps. We evaluate representative models from both open-source and closed-source. Open-source MLLMs include LLaVA-3.2-11B-Vision-Instruct~\cite{llava32vision}, LLaVA-3.2-90B-Vision-Instruct~\cite{llava32vision}, Qwen2.5-VL-32B-Instruct~\cite{qwen25vl32b}, Qwen2.5-VL-72B-Instruct~\cite{qwen25vl72b}, and Qwen-VL-max~\cite{qwenvlmax}. Closed-source MLLMs include Gemini-1.5 Flash~\cite{gemini15flash}, GPT-4 Vision~\cite{gpt4vision}, and LLaMA-4-Maverick~\cite{llama4maverick}, recognized for their sophisticated multi-modal fusion and retrieval-based response capabilities.

\begin{wraptable}{r}{0.6\linewidth}
\centering
\vspace{-18pt}
\caption{Comparison of representative multi-modal reasoning datasets. \textbf{V} (Visual Input), \textbf{OC} (Optical Characters), \textbf{I+T} (Image + Text), \textbf{IL} (Interleaved Format), \textbf{TJ} (Thinking Judgment) and \textbf{Source} (W: Web, T: Textbook, R: Remake).}
\label{tab:mmmr_comparison}
\resizebox{0.6\textwidth}{!}{
\begin{tabular}{lcccccc}
\toprule
\textbf{Dataset} & \textbf{Size} & \textbf{Images} & \textbf{Format} & \textbf{Source} & \textbf{Reason} & \textbf{TJ} \\
\midrule
VQA~\cite{antol2015vqa}        & >1M   & V        & I+T      & W      & Low     & \faTimesCircle[regular] \\
GQA~\cite{hudson2019gqa}        & >1M   & V        & I+T      & R      & Medium  & \faTimesCircle[regular] \\
VizWiz~\cite{bigham2010vizwiz}
     & 32K   & V        & I+T      & W      & Low     & \faTimesCircle[regular] \\
TextVQA~\cite{textvqa}    & 45K   & OC       & I+T      & W      & Medium  & \faTimesCircle[regular] \\
OK-VQA~\cite{marino2019ok}      & 14K   & V+OC     & I+T      & W      & Medium  & \faTimesCircle[regular] \\
SEED~\cite{li2024seed}       & 19K   & V+OC     & I+T      & W      & Medium  & \faTimesCircle[regular] \\
MMBench~\cite{liu2023mmbench}    & 3K    & V+OC     & I+T      & W+R      & Medium  & \faTimesCircle[regular] \\
MM-Vet~\cite{yu2023mmvet}     & 0.2K  & V+OC     & I+T      & W      & Medium  & \faTimesCircle[regular] \\
ScienceQA~\cite{lu2022Science}  & 6K    & 5 Types  & I+T      & T      & Medium  & \faTimesCircle[regular] \\
MME-COT~\cite{mmecot}    & 1.1K  & -        & IL       & W+R      & Medium  & \faTimesCircle[regular] \\
EMMA~\cite{emma}       & 2.7K  & -        & IL       & W+R      & Medium  & \faTimesCircle[regular] \\
MMMU~\cite{mmmu}       & 11.5K & 30 Types & IL       & W+T    & Medium  & \faTimesCircle[regular] \\
\midrule
\MMMR & 1.1K & 15 Types & IL & \textbf{W+T+R} & \textbf{High} & \faCheckCircle \\
\bottomrule
\end{tabular}
}
\vspace{-10pt}
\end{wraptable}

\noindent\textbf{Multi-Modal Language Models with Thinking (MLLMs-T).} MLLMs-T extend the capabilities of MLLMs by producing intermediate reasoning traces before final answer generation. This architecture allows structured, step-wise reasoning and facilitates deeper interpretability in complex problem-solving. We include open models such as QVQ-72B-Preview~\cite{qvq72bpreview}, and several advanced proprietary models, including Gemini-2.0 Flash~\cite{gemini20flash}, Gemini-2.5 Pro~\cite{gemini25pro}, Claude-3.7-sonnet~\cite{claude37sonnet}, and o4-mini~\cite{o4mini}.
A particularly notable configuration is our custom \emph{Dual Model}, which simulates MLLMs-T behavior by integrating the strengths of two distinct models. Specifically, GPT-4V~\cite{gpt4vision} is responsible for parsing the input question and image content—handling rich visual understanding and language grounding. The parsed task is then passed to DeepSeek-R1~\cite{deepseekr1} for structured multi-step reasoning. This pipeline allows us to isolate and examine the effect of high-quality Thinking traces independently of perception noise. Moreover, since DeepSeek-R1 is optimized for textual reasoning rather than vision, this dual formulation ensures modular design and enhanced control over each reasoning stage. 

\begin{figure}[!t]
\centering
\includegraphics[width=1\textwidth]{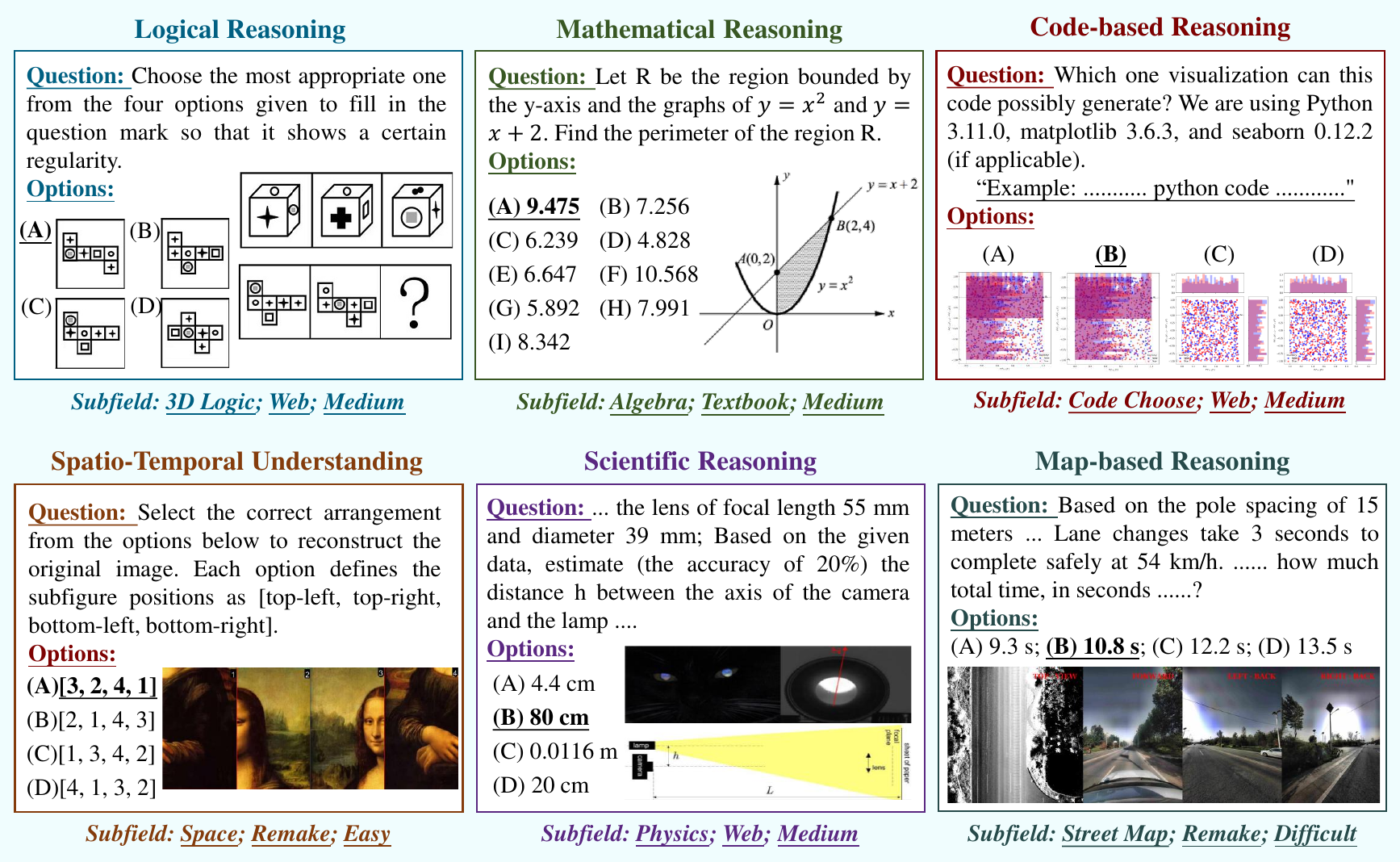}
\caption{Representative multi-modal reasoning samples from \MMMR. Each example consists of interleaved image-text input, a complex reasoning question, annotation, and source information. The samples illustrate the dataset’s high structural variability and reasoning depth.}
\label{fig:data}
\vspace{-20pt}
\end{figure}

\begin{wraptable}{r}{0.48\textwidth}
\vspace{-35pt}
\caption{Statistics of \MMMR, detailing the distribution and characteristics of six tasks.
}
\label{tab:dataset_stats}
\renewcommand{\arraystretch}{1}
\resizebox{0.49\textwidth}{!}{
\begin{tabular}{l r}
\toprule
\textbf{Category} & \textbf{Value} \\
\midrule
\rowcolor{gray!10} \multicolumn{2}{c}{\textbf{Quantitative statistics}} \\
Total Questions & 1083 \\
Total Subjects/Subfields & 6/16 \\
Image Types & 15 \\ \midrule
\rowcolor{gray!10} \multicolumn{2}{c}{\textbf{Dataset Split}} \\
Validation:Test & 106:977 \\
Difficulties (Easy:Medium:Hard) & 30\%:40\%:30\% \\ \midrule
\rowcolor{gray!10} \multicolumn{2}{c}{\textbf{Task Type Distribution}} \\
Logical Reasoning & 182 (16.8\%) \\
Mathematical Reasoning & 212 (19.5\%) \\
Spatio-Temporal Understanding & 200 (18.5\%) \\
Code Reasoning & 141 (13.0\%) \\
Map Reasoning & 150 (13.8\%) \\
Science Reasoning & 198 (18.4\%) \\ \midrule
\rowcolor{gray!10} \multicolumn{2}{c}{\textbf{Content Characteristics}} \\
Average Question Length & 68.75 words \\
Average Option Length & 12.25 words \\
Average Response Length & 135.60 words \\
Multi-modal Inputs per Question & 1.85 \\
Reasoning Steps per Question & 3.42 \\ \midrule
\rowcolor{gray!10} \multicolumn{2}{c}{\textbf{Annotation and Complexity}} \\
Remade Questions & 483 (44.6\%) \\
Average Reasoning Depth & 4.15 levels \\
Cross-Modal Integration Rate & 95.2\% \\
\bottomrule
\end{tabular}
}
\vspace{-20pt}
\end{wraptable}

\noindent\textbf{Datasets.}~~The \MMMR~is designed from first principles to meet the demands of benchmarking multi-modal reasoning with Thinking.Compared to prior works like MMMU~\cite{mmmu}, MME-CoT~\cite{mmecot}, and EMMA~\cite{emma}, \MMMR~provides a full-spectrum redesign of reasoning task settings. Each of its 1,083 problems is carefully constructed and categorized into six reasoning types and sixteen fine-grained subfields, providing targeted coverage of logical, mathematical, spatio-temporal, code, map-based, and scientific reasoning. 
The dataset incorporates a rich array of visual stimuli, including charts, diagrams, 3D maps, and visual code logic, covering 15 unique image types. Unlike retrieval-centric tasks, many questions require long-horizon reasoning, abstraction, or visual-spatial synthesis. To further increase complexity, 44.6\% of the items are remade or enhanced beyond web or textbook sources.
Moreover, \MMMR~facilitates intermediate reasoning evaluation. For each sample, the source origin (Web, Textbook, Remake), and task type are documented, supporting both output-based and process-based analysis. This structure enables deep introspection into where and how reasoning fails or succeeds, making it a unique testbed for evaluating MLLMs-T.

\noindent\textbf{Metrics.}~~We evaluate MLLMs-T and MLLMs on the \MMMR~using a concise suite of metrics, with scores normalized to [0, 1]: 1) ACC, the proportion of correct answers; 2) RTQ, assessing how well the Thinking process aligns with the problem’s requirements, independent of answer correctness; 3) RTA, evaluating the logical consistency between the thinking process and the final answer, regardless of accuracy; 4) Reasoning Step Consistency (RSC), measuring logical coherence across Thinking steps through consistency checks. 

  \section{Empirical Results and Analysis}\label{sec:experiments}

\subsection{Main Results}

\begin{table}[t]
\centering
\caption{Accuracy (\%) comparison of baselines, MLLMs, and MLLMs-T on the \MMMR. Each row highlights the per-model highest and lowest scores using \colorgreen{green} and \colorred{red}, respectively. For each column (task type), the best-performing model is indicated in \textbf{bold} and the second-best is \underline{underline}. Models marked with * are closed-source. ``S-T'' denotes the Space-Time.}
\label{tab:main_results}
\renewcommand{\arraystretch}{1}
\resizebox{\textwidth}{!}{
\begin{tabular}{l c c c c c c c c}
\toprule
 & \textbf{Validation} & \textbf{Test} & \textbf{Logic} & \textbf{Math} & \textbf{S-T} & \textbf{Code} & \textbf{Map} & \textbf{Science} \\
 & \textbf{(106)} & \textbf{(977)} & \textbf{(182)} & \textbf{(212)} & \textbf{(200)} & \textbf{(141)} & \textbf{(150)} & \textbf{(198)} \\
\midrule
\rowcolor{gray!10}
\multicolumn{9}{c}{\textbf{Baselines}} \\
Random Choice & 22.1 & 23.62 & 24.18 & 24.06 & 21.50 & 25.53 & 22.67 & 23.74 \\
Frequent Choice & 26.8 & 26.58 & 26.92 & 26.42 & 24.00 & 24.82 & 25.33 & 29.80 \\
Expert (Human only) & 29.23 & - & - & - & - & - & - & - \\
Expert (Human + GPT-4o~\cite{hurst2024gpt}) & \underline{\textbf{52.85}} & - & - & - & - & - & - & - \\
\midrule
\rowcolor{gray!10}
\multicolumn{9}{c}{\textbf{Multi-Modal Large Language Models without Thinking}} \\
LLaVA-3.2-11B-Vision-Instruct~\cite{llava32vision} & 24.53 & 23.92 & 18.68 & \colorgreen{31.13} & 28.00 & \colorred{13.48} & 22.67 & 22.73 \\
LLaVA-3.2-90B-Vision-Instruct~\cite{llava32vision} & 30.19 & 27.65 & 21.43 & 34.91 & \colorgreen{35.00} & \colorred{17.73} & 25.33 & 21.72 \\
Qwen2.5-VL-32B-Instruct~\cite{qwen25vl32b} & 34.86 & 34.90 & 25.27 & \colorgreen{45.28} & 45.00 & 32.62 & 36.67 & \colorred{21.72} \\
Qwen2.5-VL-72B-Instruct~\cite{qwen25vl72b} & 36.95 & 37.18 & \colorred{24.18} & 46.70 & \colorgreen{47.50} & \textbf{41.84} & \textbf{42.67} & 31.31 \\
Qwen-VL-max~\cite{qwenvlmax} & 35.13 & 35.55 & \colorred{24.18} & \colorgreen{47.17} & 46.00 & 39.01 & 35.33 & 28.28 \\
Gemma-3-27B-IT~\cite{gemma3report2025} & 30.87 & 29.01 & \colorred{22.53} & \colorgreen{42.45} & 33.50 & 34.75 & 26.67 & 27.27 \\
Gemini-1.5 Flash*~\cite{gemini15flash} & 32.18 & 29.61 & 28.57 & \colorgreen{37.74} & 37.00 & \colorred{18.44} & 24.67 & 32.83 \\
GPT-4 Vision*~\cite{gpt4vision} & 37.59 & 38.05 & \colorred{28.02} & 35.85 & \colorgreen{49.00} & 28.37 & 32.00 & 41.92 \\
LLaMA-4-Maverick*~\cite{llama4maverick} & 40.68 & \underline{41.82} & 30.77 & 44.81 & \colorgreen{46.00} & \underline{37.59} & \colorred{30.67} & 38.38 \\
\midrule
\rowcolor{gray!10}
\multicolumn{9}{c}{\textbf{Multi-Modal Large Language Models with Thinking}} \\
QVQ-72B-Preview~\cite{qvq72bpreview} & 30.94 & 32.09 & \colorred{26.37} & 38.21 & \colorgreen{42.00} & 32.62 & 31.33 & 32.83 \\
Gemini-2.0 Flash*~\cite{gemini20flash} & 37.63 & 37.89 & 35.16 & \colorgreen{\textbf{50.47}} & \underline{49.50} & \colorred{28.37} & 30.67 & 41.41 \\
Gemini-2.5 Pro*~\cite{gemini25pro} & \textbf{42.45} & \textbf{42.36} & \textbf{39.56} & 41.51 & 44.50 & \colorred{36.17} & \underline{37.33} & \colorgreen{\textbf{46.46}} \\
Claude-3.7-sonnet*~\cite{claude37sonnet} & 38.28 & 37.72 & \underline{35.71} & 45.75 & \colorgreen{\textbf{51.00}} & \colorred{21.28} & 34.00 & 43.94 \\
o4-mini*~\cite{o4mini} & 38.64 & 37.58 & 34.62 & 46.23 & \colorgreen{47.50} & \colorred{19.86} & 29.33 & 41.41 \\
Dual (\cite{gpt4vision} + DeepSeek-R1~\cite{deepseekr1}) & \underline{41.26} & 41.00 & \underline{35.71} & \underline{47.64} & \colorgreen{48.00} & 22.70 & \colorred{22.67} & \underline{45.45} \\
\bottomrule
\end{tabular}
}
\vspace{-20pt}
\end{table}

We evaluate 17 models on the \MMMR, where baselines include \emph{Random Choice} and \emph{Frequent Choice}, which serve as naïve heuristics, and two \emph{Expert} configurations that represent human upper bounds with or without model assistance (see Appendix~\ref{expert} for full descriptions). As shown in Table~\ref{tab:main_results}, MLLMs-T overall outperform MLLMs across six tasks, highlighting the advantage of incorporating explicit thinking mechanisms. Notably, \textit{Gemini-2.5 Pro} achieves the highest overall test accuracy at 42.36\%, while the \textit{Expert (Human + GPT-4o)} attains an upper-bound of 52.85\%, indicating a remaining gap between state-of-the-art MLLMs-T and human-assisted reasoning [RQ1 Summary].

\textbf{Model-wise performance reflects generalization differences.}  
By examining the best and second-best performers for each task column, we observe that \textit{Gemini-2.5 Pro} (with 4 best and 1 second-best scores) exhibits the most stable and competitive accuracy across reasoning types. This suggests that explicit reasoning modules, when paired with carefully supervised thinking strategies, enable strong generalization across diverse task formats. \textit{Gemini-2.0 Flash} performs robustly in Math (50.47\%) but shows a significant drop in Code, indicating limited cross-domain transferability. The \textit{Dual} achieves strong results in Logic and Science, supporting the effectiveness of modular architectural design. In contrast, open-source models like \textit{Qwen2.5-VL-72B} and \textit{Qwen-VL-max} display isolated strengths in domains such as Spatio-Temporal reasoning and Code, but lack consistency. Weaker models (e.g., \textit{Gemma-3-27B} and \textit{LLaVA-3.2-11B}) show broad performance variance and struggle particularly with symbolically dense or spatial tasks, indicating limitations in reasoning depth despite model size.

\textbf{Task-wise analysis reveals variation in reasoning difficulty.}  
A row-wise comparison of model accuracy extremes reveals task-dependent performance variability. \textit{Math} and \textit{Space-Time} tasks, which dominate the best-performing entries (green highlights), are generally more tractable, suggesting progress in symbolic computation and spatial comprehension. In contrast, tasks like \textit{Logic} and especially \textit{Code} show lower accuracy ceilings (often below 42\%) and greater inter-model dispersion, as evidenced by the concentration of minimum scores (red highlights). These patterns underscore the utility of \MMMR~in enabling fine-grained analysis of multimodal reasoning capabilities, offering a more diagnostic lens than aggregate performance alone.


\subsection{Thinking Quality Analysis}\label{sec:rtep}

\begin{figure}[t]
\centering
\includegraphics[width=0.9\textwidth]{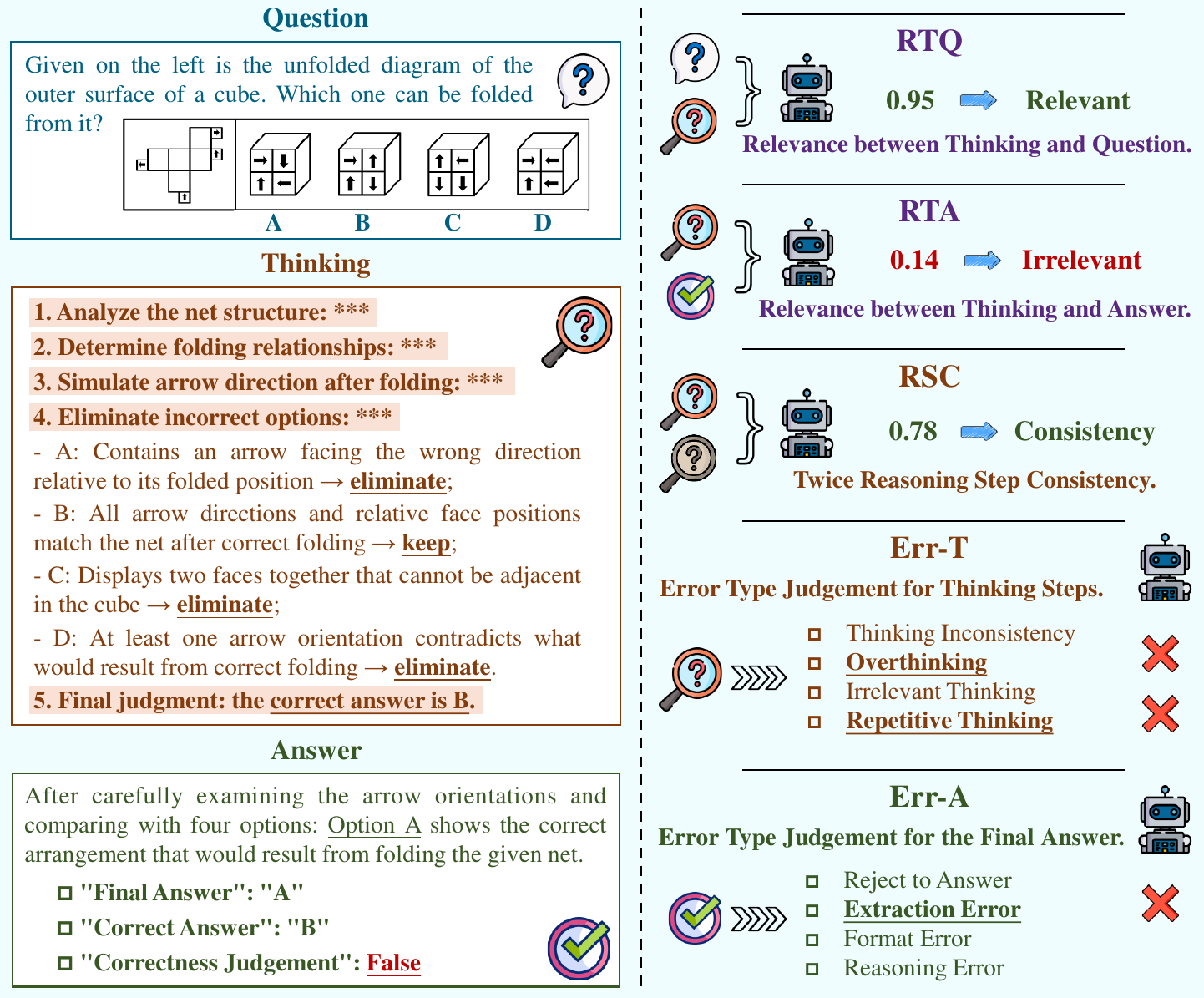}
\caption{Overview of the Reasoning Trace Evaluation Pipeline (RTEP). The pipeline applies structured scoring of intermediate reasoning traces, evaluating consistency, relevance, and verbosity.}
\vspace{-10pt}
\label{fig:thinking}
\end{figure}

To evaluate intermediate reasoning quality beyond answer correctness, we introduce the Reasoning Trace Evaluation Pipline (RTEP). This structured pipeline annotates and scores each model's reasoning trace across three dimensions: Relevance to the Question (RTQ), Relevance to the Answer (RTA), and Reasoning Step Consistency (RSC), each rated on a 0--10 scale. Leveraging GPT-4o~\cite{hurst2024gpt} as an automated evaluator, RTEP enables scalable, semantically aligned trace assessments, avoiding the subjectivity and cost of manual annotation. As illustrated in Figure~\ref{fig:thinking}, this design allows for model-agnostic diagnosis of coherence, verbosity, and reasoning alignment, enabling targeted comparisons across architectures and task types.

Table~\ref{tab:thinking_quality} summarizes a detailed comparison between Claude-3.7-sonnet and Dual (GPT-4V + DeepSeek-R1) across all six reasoning tasks in \MMMR. Claude-3.7 consistently outperforms Dual in Overall Score (OS), with especially strong results in Math and Science, where compact, logically consistent traces are essential. In contrast, the Dual system achieves marginally higher answer accuracy in several tasks (e.g., Logic: +2.79\%), but at the cost of reasoning coherence, as reflected in lower OS values and significantly inflated trace lengths (TLen often 3--5× higher). This indicates that longer outputs, while occasionally improving accuracy, tend to dilute reasoning relevance and introduce redundancy or inconsistency. For instance, in Code and Space-Time tasks, Dual's modular pipeline leads to repetitive or loosely linked steps, revealing that increased trace length does not ensure better reasoning quality. These findings emphasize that answer correctness alone is insufficient as a proxy for reasoning performance—models with higher accuracy can still produce flawed, verbose, or semantically incoherent rationales. Evaluating trace quality is thus essential for advancing the robustness and interpretability of MLLMs-T.

Overall, accurate answers do not guarantee sound reasoning. Our findings reveal that high-performing MLLMs-T can still produce incoherent thought processes, suggesting that future progress hinges not only on output correctness but on the quality of the reasoning path itself [RQ2 Summary].

\begin{table}[t]
\centering
\caption{Comparison of reasoning quality between Claude-3.7-sonnet and Dual across six task types. RTQ, RTA, RSC are reasoning trace metrics in [0--10]; ACC is final answer accuracy (\%); OS is a weighted overall score (\textit{0.3$\cdot$RTQ + 0.3$\cdot$RTA + 0.3$\cdot$RSC + 0.1$\cdot$(ACC$\times$0.1)}); TLen denotes trace length in thousands of tokens; ThinkErr indicates dominant reasoning flaw (defined in Section~\ref{sec:thinking error}).}
\label{tab:thinking_quality}
\renewcommand{\arraystretch}{1}
\resizebox{0.98\textwidth}{!}{
\begin{tabular}{l c |>{\columncolor{gray!5}}c >{\columncolor{gray!5}}c >{\columncolor{gray!5}}c 
|>{\columncolor{yellow!2}}>{\centering\arraybackslash}m{1.6cm}
|>{\columncolor{yellow!2}}>{\centering\arraybackslash}m{1.6cm}
|>{\columncolor{yellow!2}}>{\centering\arraybackslash}m{1.6cm}
|>{\centering\arraybackslash}m{3cm}}
\toprule
\textbf{Task} & \textbf{Model} & \textbf{RTQ} & \textbf{RTA} & \textbf{RSC} 
& \textbf{ACC (\%)} & \textbf{OS} & \textbf{TLen (k)} & \textbf{ThinkErr} \\
\midrule

\multirow{3}{*}{\textbf{Logic}}
  & Claude-3.7-sonnet & 9.39 & 9.41 & 9.07 & 35.71 & 8.72 & 3.71 & Overthinking \\
  & Dual              & 6.32 & 7.63 & 6.18 & 38.50 & 6.42 & 15.19 & Irrelevant Thinking \\
  & $\Delta$ (Dual−Claude) &  -  &  -  &   -  & \colorgreen{+ 2.79} & \colorred{− 2.30} & \colorgreen{+ 11.48} & - \\
\midrule

\multirow{3}{*}{\textbf{Math}}
  & Claude-3.7-sonnet & 8.88 & 9.02 & 8.40 & 45.75 & 8.35 & 5.32 & Overthinking \\
  & Dual              & 8.57 & 8.80 & 7.82 & 47.60 & 8.03 & 21.35 & Overthinking \\
  & $\Delta$ (Dual−Claude) &  -  &  -  &   -   & \colorgreen{+ 1.85} & \colorred{− 0.32} & \colorgreen{+ 16.03} & - \\
\midrule

\multirow{3}{*}{\textbf{Space-Time}}
  & Claude-3.7-sonnet & 9.50 & 9.26 & 8.97 & 51.00 & 8.83 & 2.38 & Overthinking \\
  & Dual              & 8.50 & 8.75 & 7.60 & 48.00 & 8.32 & 14.83 & Repetitive Thinking \\
  & $\Delta$ (Dual−Claude) &   -  &  -  &   -   & \colorred{− 3.00} & \colorred{− 0.51} & \colorgreen{+ 12.45} & - \\
\midrule

\multirow{3}{*}{\textbf{Code}}
  & Claude-3.7-sonnet & 9.56 & 9.31 & 9.30 & 21.28 & 8.82 & 4.53 & Repetitive Thinking \\
  & Dual              & 8.61 & 8.56 & 7.94 & 22.90 & 7.93 & 17.24 & Inconsistency \\
  & $\Delta$ (Dual−Claude) &  -  &  -  &   -   & \colorgreen{+ 1.62} & \colorred{− 0.89} & \colorgreen{+ 12.71} & - \\
\midrule

\multirow{3}{*}{\textbf{Map}}
  & Claude-3.7-sonnet & 9.17 & 8.94 & 8.43 & 23.80 & 8.57 & 1.76 & Irrelevant Thinking \\
  & Dual              & 7.08 & 7.42 & 6.42 & 22.50 & 6.99 & 12.43 & Repetitive Thinking \\
  & $\Delta$ (Dual−Claude) &   -  &  -  &   -   & \colorred{− 1.30} & \colorred{− 1.58} & \colorgreen{+ 10.67} & - \\
\midrule

\multirow{3}{*}{\textbf{Science}}
  & Claude-3.7-sonnet & 8.95 & 9.29 & 8.73 & 43.93 & 8.77 & 3.61 & Inconsistency \\
  & Dual              & 8.25 & 8.84 & 7.81 & 45.10 & 8.32 & 14.50 & Inconsistency \\
  & $\Delta$ (Dual−Claude) &  -  &  -  &   -    & \colorgreen{+ 1.17} & \colorred{− 0.45} & \colorgreen{+ 10.89} & - \\
\bottomrule
\end{tabular}
}
\vspace{-15pt}
\end{table}

\subsection{Thinking and Answer Error Types Analysis}\label{sec:thinking error}

\begin{wrapfigure}{r}{0.4\textwidth}
\centering
\includegraphics[width=0.4\textwidth]{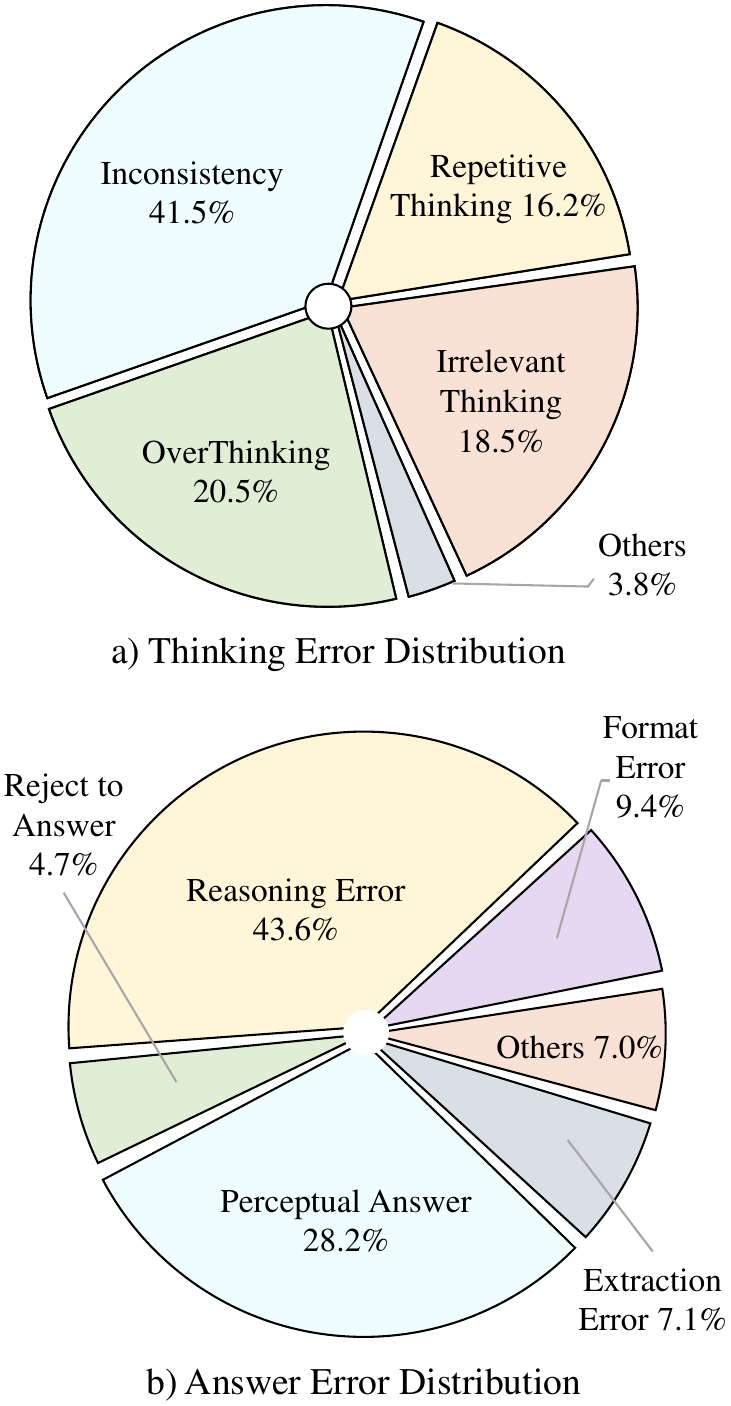}
\caption{Distribution of Thinking and Answer Errors on Claude-3.7-sonnet.}
\label{fig:error_types_combined}
\vspace{-45pt}
\end{wrapfigure}

To understand the structural weaknesses in multimodal reasoning, we analyze errors in both intermediate \emph{Thinking} traces and final \emph{Answer} predictions of Claude-3.7-sonnet on the \MMMR~validation set. As visualized in Figure~\ref{fig:error_types_combined}, we classify each error into semantically distinct categories, allowing targeted diagnosis of reasoning failures.

\textbf{Thinking Errors Distribution.}

\underline{\textit{1) Inconsistency (41.5\%)}} reflects internal contradictions or self-conflicting logic, often arising in Science or Logic tasks where maintaining state across steps is nontrivial.

\underline{\textit{2) Overthinking (20.5\%)}} denotes unnecessarily verbose or speculative reasoning paths. These are prevalent in otherwise simple tasks where compact reasoning suffices.

\underline{\textit{3) Irrelevant Thinking (18.5\%)}} includes content unrelated to the question or answer. These errors typically occur in poorly grounded inputs or under weak alignment.

\underline{\textit{4) Repetitive Thinking (16.2\%)}} captures duplication without informational gain, frequently observed in Code and Map, where step-tracking or termination is difficult.

\underline{\textit{5) Others (3.8\%)}} contain rare phenomena such as speculative completion or omitted critical steps.

\textbf{Answer Errors Distribution.}

\underline{\textit{1) Reasoning Error (43.6\%)}} indicates logically flawed reasoning that nonetheless produces confident but incorrect answers, especially common in Math and Science.

\underline{\textit{2) Perceptual Error (28.2\%)}} reflects misinterpretation of visual data such as spatial layouts or charts—frequent in Map and Space-Time tasks.

\underline{\textit{3) Format Error (9.4\%)}} denotes violations of expected output formats, such as missing labels or extraneous text, often due to instruction-following deficiencies.  

\underline{\textit{4) Answer Extraction Error (7.1\%)}} occurs when models generate lengthy thinking traces but omit or fail to commit to a final answer—highlighting uncertainty or incomplete reasoning convergence.

\underline{\textit{5) Reject to Answer (4.7\%)}} involves abstention despite solvable inputs, typically due to cautious decoding or alignment penalties.

\underline{\textit{6) Others (7.0\%)}} include ambiguous completions or partially correct statements.

\textbf{Error Analysis.}~~The error distributions suggest that high answer accuracy often masks underlying reasoning path defects. The dominance of inconsistency and overthinking in reasoning traces reveals fundamental challenges in maintaining logical control and brevity. Likewise, the prevalence of reasoning-based answer errors over perceptual ones underscores that symbolic structure, rather than visual understanding, remains the primary bottleneck in high-level multimodal cognition. These findings reinforce the importance of trace-aware evaluation: coarse answer-level metrics alone cannot capture reasoning fidelity [RQ3 Summary].

  \section{Related Work}
\label{sec:related}

\textbf{Multi-Modal Large Language Models Benchmarking.}~~The evaluation of multimodal reasoning has progressed from VQAv2~\cite{antol2015vqa}, GQA~\cite{hudson2019gqa}, and VCR~\cite{zellers2019vcr} to broader and more specialized datasets such as TextVQA~\cite{textvqa}, ScienceQA~\cite{lu2022Science}, AI2D~\cite{ai2d}, and SEED~\cite{li2024seed}, and recent large-scale benchmarks like MMBench~\cite{liu2023mmbench}, MM-Vet~\cite{yu2023mmvet}, EMMA~\cite{emma}, MathVista~\cite{lu2024mathvista}, and MMMU~\cite{mmmu}. These datasets have advanced task diversity and domain specificity, yet most focus primarily on answer accuracy with limited insight into reasoning quality. While efforts such as MME-CoT~\cite{mmecot} introduce reasoning trace annotations, they mainly rely on additional CoT designs. In contrast, our \MMMR~is constructed as a high-difficulty, multi-domain dataset specifically for evaluating multimodal reasoning. It not only spans six distinct reasoning types with structured task design but also supports fine-grained assessment of thinking in MLLMs-T, offering a comprehensive diagnostic standard for future multimodal models.

\textbf{Reasoning Traces and Thinking Evaluation.}~~In textual LLMs, reasoning trace prompting methods like Chain-of-Thought~\cite{wei2022chain}, ReAct~\cite{yao2022react}, and Reflexion~\cite{shinn2023reflexion} have improved interpretability and performance through explicit step-by-step reasoning. Recent works propose evaluation tools such as RATER~\cite{liu2023rater} and DECKARD~\cite{liu2023deckard} to assess coherence, faithfulness, and hallucination in these traces. However, such evaluations are still limited to language-only settings. Our work fills this gap by enabling trace-level reasoning evaluation within a multimodal benchmark, supporting both process- and outcome-oriented assessments.
  
  \section{Conclusion}\label{sec:conclusion}
This paper presents \MMMR, a new benchmark and evaluation framework for advancing the study of multi-modal reasoning in large language models. Distinct from prior efforts that primarily emphasize perception or answer correctness, \MMMR~targets high-complexity, symbolic reasoning across six diverse domains, including logic, mathematics, and space-time inference. To systematically assess reasoning fidelity, we propose the Reasoning Trace Evaluation Pipeline (RTEP), which incorporates structured metrics (RTQ, RTA, RSC), length-efficiency analysis, and error-type classification to evaluate the coherence and relevance of intermediate thinking. Through extensive experiments on 17 models, we find that MLLMs-T overall outperform standard MLLMs in tasks requiring structured reasoning. Our findings suggest that improving multi-modal reasoning requires not just stronger instruction tuning or scale, but more cognitively aligned architectures that optimize for both answer correctness and thinking quality. We hope this benchmark catalyzes further research on reflective reasoning, modular cognition, and generalizable multi-modal understanding.

\textbf{Limitations.}~~While \MMMR~emphasizes reasoning difficulty and multi-modal integration, it does not explicitly define fine-grained difficulty levels or hierarchical task groupings. This limits the granularity of comparative analysis across reasoning complexity levels. The main challenge lies in accurately quantifying reasoning difficulty across diverse modalities and task structures, which requires both task-specific cognitive modeling and robust annotation protocols. Future work should explore dynamic task stratification to better support curriculum learning, diagnostic evaluation, and model scalability studies.\label{sec:limitation}

  \bibliographystyle{plain}
  \bibliography{neurips_2025}

\begin{thebibliography}{10}

\bibitem{qwenvlmax}
Alibaba~DAMO Academy.
\newblock Qwen-vl-max: Most capable visual-language model.
\newblock \url{https://github.com/QwenLM/Qwen-VL}, 2024.
\newblock Accessed: 2025-05-14.

\bibitem{qwen25vl32b}
Alibaba~DAMO Academy.
\newblock Qwen2.5-vl-32b-instruct: Vision-language model.
\newblock \url{https://huggingface.co/Qwen/Qwen2.5-VL-32B-Instruct}, 2025.
\newblock Accessed: 2025-05-14.

\bibitem{qwen25vl72b}
Alibaba~DAMO Academy.
\newblock Qwen2.5-vl-72b-instruct: Vision-language model.
\newblock \url{https://huggingface.co/Qwen/Qwen2.5-VL-72B-Instruct}, 2025.
\newblock Accessed: 2025-05-14.

\bibitem{llama4maverick}
Meta AI.
\newblock Llama 4 maverick: Natively multimodal model.
\newblock \url{https://ai.meta.com/blog/llama-4-multimodal-intelligence/}, 2025.
\newblock Accessed: 2025-05-14.

\bibitem{qvq72bpreview}
Qwen AI.
\newblock Qvq-72b-preview: Vision-language model.
\newblock \url{https://huggingface.co/Qwen/QVQ-72B-Preview}, 2025.
\newblock Accessed: 2025-05-14.

\bibitem{claude37sonnet}
Anthropic.
\newblock Claude 3.7 sonnet: Multimodal model.
\newblock \url{https://www.anthropic.com/claude-3-7-sonnet}, 2025.
\newblock Accessed: 2025-05-14.

\bibitem{antol2015vqa}
Stanislaw Antol, Aishwarya Agrawal, Jiasen Lu, Margaret Mitchell, Dhruv Batra, C~Lawrence~Zitnick, and Devi Parikh.
\newblock Vqa: Visual question answering.
\newblock In {\em Proceedings of the IEEE international conference on computer vision}, pages 2425--2433, 2015.

\bibitem{bai2023qwen}
Jinze Bai, Shuai Bai, Shusheng Yang, Shijie Wang, Sinan Tan, Peng Wang, Junyang Lin, Chang Zhou, and Jingren Zhou.
\newblock Qwen-vl: A versatile vision-language model for understanding, localization, text reading, and beyond.
\newblock {\em arXiv preprint arXiv:2308.12966}, 2023.

\bibitem{bigham2010vizwiz}
Jeffrey~P. Bigham, Chandrika Jayant, Hanjie Ji, Greg Little, Andrew Miller, Robert~C. Miller, Robin Miller, Aubrey Tatarowicz, Brandyn White, Samuel White, and Tom Yeh.
\newblock Vizwiz: Nearly real-time answers to visual questions.
\newblock In {\em Proceedings of the 23rd Annual ACM Symposium on User Interface Software and Technology (UIST)}, pages 333--342. ACM, 2010.

\bibitem{gemini15flash}
Google DeepMind.
\newblock Gemini 1.5 flash: Multimodal model.
\newblock \url{https://cloud.google.com/vertex-ai/generative-ai/docs/models/gemini/1-5-flash}, 2024.
\newblock Accessed: 2025-05-14.

\bibitem{gemini20flash}
Google DeepMind.
\newblock Gemini 2.0 flash: Multimodal model.
\newblock \url{https://cloud.google.com/vertex-ai/generative-ai/docs/models/gemini/2-0-flash}, 2024.
\newblock Accessed: 2025-05-14.

\bibitem{google2025gemini25}
Google DeepMind.
\newblock Gemini 2.5: Our most intelligent ai model.
\newblock \url{https://blog.google/technology/google-deepmind/gemini-model-thinking-updates-march-2025/}, 2025.
\newblock Accessed: 2025-05-14.

\bibitem{gemini25pro}
Google DeepMind.
\newblock Gemini 2.5 pro: Multimodal model.
\newblock \url{https://cloud.google.com/vertex-ai/generative-ai/docs/models/gemini/2-5-pro}, 2025.
\newblock Accessed: 2025-05-14.

\bibitem{deepseekr1}
Daya Guo, Dejian Yang, Haowei Zhang, Junxiao Song, et~al.
\newblock Deepseek-r1: Incentivizing reasoning capability in llms via reinforcement learning.
\newblock \url{https://arxiv.org/abs/2501.12948}, 2025.
\newblock Accessed: 2025-05-14.

\bibitem{emma}
Yunzhuo Hao, Jiawei Gu, Huichen~Will Wang, Linjie Li, Zhengyuan Yang, Lijuan Wang, and Yu~Cheng.
\newblock Can mllms reason in multimodality? emma: An enhanced multimodal reasoning benchmark.
\newblock \url{https://arxiv.org/abs/2501.05444}, 2025.
\newblock Accessed: 2025-05-14.

\bibitem{hudson2019gqa}
Drew~A Hudson and Christopher~D Manning.
\newblock Gqa: A new dataset for real-world visual reasoning and compositional question answering.
\newblock In {\em Proceedings of the IEEE Conference on Computer Vision and Pattern Recognition}, pages 6700--6709, 2019.

\bibitem{hurst2024gpt}
Aaron Hurst, Adam Lerer, Adam~P Goucher, Adam Perelman, Aditya Ramesh, Aidan Clark, AJ~Ostrow, Akila Welihinda, Alan Hayes, Alec Radford, et~al.
\newblock Gpt-4o system card.
\newblock {\em arXiv preprint arXiv:2410.21276}, 2024.

\bibitem{mmecot}
Dongzhi Jiang, Renrui Zhang, Ziyu Guo, Yanwei Li, Yu~Qi, Xinyan Chen, Liuhui Wang, Jianhan Jin, Claire Guo, Shen Yan, Bo~Zhang, Chaoyou Fu, Peng Gao, and Hongsheng Li.
\newblock Mme-cot: Benchmarking chain-of-thought in large multimodal models for reasoning quality, robustness, and efficiency.
\newblock \url{https://arxiv.org/abs/2502.09621}, 2025.
\newblock Accessed: 2025-05-14.

\bibitem{johnson2017clevr}
Justin Johnson, Bharath Hariharan, Laurens van~der Maaten, Li~Fei-Fei, C~Lawrence Zitnick, and Ross Girshick.
\newblock Clevr: A diagnostic dataset for compositional language and elementary visual reasoning.
\newblock In {\em Proceedings of the IEEE conference on computer vision and pattern recognition}, pages 2901--2910, 2017.

\bibitem{ai2d}
Aniruddha Kembhavi, M.~Salvato, M.~Seo, H.~Hajishirzi, and A.~Farhadi.
\newblock A diagram is worth a dozen images.
\newblock In {\em Proceedings of the European Conference on Computer Vision (ECCV)}, pages 235--251, 2016.

\bibitem{li2024mmcode}
Kaixin Li, Yuchen Tian, Qisheng Hu, Ziyang Luo, Zhiyong Huang, and Jing Ma.
\newblock Mmcode: Benchmarking multimodal large language models for code generation with visually rich programming problems.
\newblock {\em arXiv preprint arXiv:2404.09486}, 2024.

\bibitem{li2024web2code}
Kaixin Li, Yuchen Tian, Qisheng Hu, Ziyang Luo, Zhiyong Huang, and Jing Ma.
\newblock Web2code: Benchmarking multimodal large language models for code generation with visually rich programming problems.
\newblock {\em Findings of the Association for Computational Linguistics: EMNLP 2024}, pages 736--783, 2024.

\bibitem{li2024seed}
Yixuan Li, Yujie Wang, Yujie Zhang, Yifan Wang, Yujie Wang, Yixuan Li, Yujie Wang, Yujie Zhang, and Yifan Wang.
\newblock Seed-bench: Benchmarking multimodal large language models.
\newblock In {\em Proceedings of the IEEE/CVF Conference on Computer Vision and Pattern Recognition (CVPR)}, 2024.

\bibitem{liu2023llava}
Haotian Liu, Chunyuan Li, Qingyang Wu, and Yong~Jae Lee.
\newblock Visual instruction tuning.
\newblock {\em arXiv preprint arXiv:2304.08485}, 2023.

\bibitem{llava32vision}
Haotian Liu, Chunyuan Li, Qingyang Wu, and Yong~Jae Lee.
\newblock Visual instruction tuning.
\newblock \url{https://arxiv.org/abs/2304.08485}, 2023.
\newblock Accessed: 2025-05-14.

\bibitem{liu2023deckard}
Jiacheng Liu, Pan Lu, Hritik Bansal, Hannaneh Hajishirzi, and Jianfeng Gao.
\newblock Deckard: Benchmarking reasoning traces in language models.
\newblock In {\em Proceedings of the 2023 Conference on Empirical Methods in Natural Language Processing}, 2023.

\bibitem{liu2023rater}
Jiacheng Liu, Pan Lu, Hritik Bansal, Hannaneh Hajishirzi, and Jianfeng Gao.
\newblock Rater: Reference-free evaluation for cot reasoning.
\newblock In {\em Proceedings of the 2023 Conference on Empirical Methods in Natural Language Processing}, 2023.

\bibitem{liu2024mapevalvisual}
Yibo Liu et~al.
\newblock Mapeval-visual: A benchmark for visual map-based planning tasks.
\newblock {\em arXiv preprint arXiv:2405.67890}, 2024.

\bibitem{liu2024multimodalselfinstruct}
Yibo Liu et~al.
\newblock Multi-modal-self-instruct: Synthesizing complex visual reasoning context using language models.
\newblock {\em arXiv preprint arXiv:2405.12345}, 2024.

\bibitem{liu2023mmbench}
Yuxin Liu, Yuxuan Zhang, Yifan Wang, Yixuan Li, Yujie Wang, Yujie Zhang, and Yifan Wang.
\newblock Mmbench: Is your multi-modal model an all-around player?
\newblock arXiv preprint arXiv:2307.06281, 2023.

\bibitem{lu2024mathvista}
Pan Lu, Hritik Bansal, Tony Xia, Jiacheng Liu, Chunyuan Li, Hannaneh Hajishirzi, Hao Cheng, Kai-Wei Chang, Michel Galley, and Jianfeng Gao.
\newblock Mathvista: Evaluating mathematical reasoning of foundation models in visual contexts.
\newblock In {\em Proceedings of the International Conference on Learning Representations (ICLR)}, 2024.

\bibitem{lu2022Science}
Pan Lu, Swaroop Mishra, Tony Xia, Liang Qiu, Kai-Wei Chang, Song-Chun Zhu, Oyvind Tafjord, Peter Clark, and Ashwin Kalyan.
\newblock Learn to explain: Multimodal reasoning via thought chains for science question answering.
\newblock In {\em Advances in Neural Information Processing Systems (NeurIPS)}, 2022.

\bibitem{lu2022learn}
Pan Lu, Xiang Wang, Zehao Lin, Zekun Zhang, Mo~Yu, Zhiyuan Yu, et~al.
\newblock Learn from peers: Equipping multi-modal learners with cross-modal memory.
\newblock In {\em Advances in Neural Information Processing Systems (NeurIPS)}, 2022.

\bibitem{marino2019ok}
Kenneth Marino, Mohammad Rastegari, Ali Farhadi, and Hannaneh Hajishirzi.
\newblock Ok-vqa: A visual question answering benchmark requiring external knowledge.
\newblock In {\em Proceedings of the IEEE/CVF Conference on Computer Vision and Pattern Recognition (CVPR)}, pages 3195--3204, 2019.

\bibitem{gpt4vision}
OpenAI.
\newblock Gpt-4 vision: Multimodal model.
\newblock \url{https://openai.com/index/gpt-4v-system-card/}, 2023.
\newblock Accessed: 2025-05-14.

\bibitem{o4mini}
OpenAI.
\newblock o4-mini: Multimodal model.
\newblock \url{https://openai.com/o4-mini}, 2025.
\newblock Accessed: 2025-05-14.

\bibitem{radford2021learning}
Alec Radford, Jong~Wook Kim, Chris Hallacy, Aditya Ramesh, Gabriel Goh, Sandhini Agarwal, Girish Sastry, Amanda Askell, Pamela Mishkin, Jack Clark, Scott Krueger, and Ilya Sutskever.
\newblock Learning transferable visual models from natural language supervision.
\newblock In {\em Proceedings of the International Conference on Machine Learning (ICML)}, pages 8748--8763, 2021.

\bibitem{sharma2018conceptual}
Piyush Sharma, Nan Ding, Sebastian Goodman, and Radu Soricut.
\newblock Conceptual captions: A cleaned, hypernymed, image alt-text dataset for automatic image captioning.
\newblock In {\em Proceedings of the 56th Annual Meeting of the Association for Computational Linguistics (Volume 1: Long Papers)}, pages 2556--2565, 2018.

\bibitem{shinn2023reflexion}
Noah Shinn, Federico Cassano, Edward Berman, Ashwin Gopinath, and Karthik Narasimhan.
\newblock Reflexion: Language agents with verbal reinforcement learning.
\newblock In {\em Proceedings of the 2023 Conference on Empirical Methods in Natural Language Processing}, 2023.

\bibitem{textvqa}
Amanpreet Singh, Vivek Natarajan, Xinlei Jiang, Xi~Chen, Marcus Rohrbach, Dhruv Batra, and Devi Parikh.
\newblock Towards vqa models that can read.
\newblock In {\em Proceedings of the IEEE/CVF Conference on Computer Vision and Pattern Recognition (CVPR)}, pages 8317--8326, 2019.

\bibitem{gemma3report2025}
Gemma Team, Aishwarya Kamath, Johan Ferret, Shreya Pathak, Nino Vieillard, Ramona Merhej, Sarah Perrin, Tatiana Matejovicova, Alexandre Ramé, Morgane Rivière, Louis Rouillard, Thomas Mesnard, Geoffrey Cideron, Jean-Bastien Grill, Sabela Ramos, Edouard Yvinec, Michelle Casbon, Etienne Pot, Ivo Penchev, Gaël Liu, Francesco Visin, Kathleen Kenealy, Lucas Beyer, Xiaohai Zhai, Anton Tsitsulin, Robert Busa-Fekete, Alex Feng, Noveen Sachdeva, Benjamin Coleman, Yi~Gao, Basil Mustafa, Iain Barr, Emilio Parisotto, David Tian, Matan Eyal, Colin Cherry, Jan-Thorsten Peter, Danila Sinopalnikov, Surya Bhupatiraju, Rishabh Agarwal, Mehran Kazemi, Dan Malkin, Ravin Kumar, David Vilar, Idan Brusilovsky, Jiaming Luo, Andreas Steiner, Abe Friesen, Abhanshu Sharma, Abheesht Sharma, Adi~Mayrav Gilady, Adrian Goedeckemeyer, Alaa Saade, Alex Feng, Alexander Kolesnikov, Alexei Bendebury, Alvin Abdagic, Amit Vadi, András György, André~Susano Pinto, Anil Das, Ankur Bapna, Antoine Miech, Antoine Yang, Antonia Paterson, Ashish
  Shenoy, Ayan Chakrabarti, Bilal Piot, Bo~Wu, Bobak Shahriari, Bryce Petrini, Charlie Chen, Charline~Le Lan, Christopher~A. Choquette-Choo, CJ~Carey, Cormac Brick, Daniel Deutsch, Danielle Eisenbud, Dee Cattle, Derek Cheng, Dimitris Paparas, Divyashree~Shivakumar Sreepathihalli, Doug Reid, Dustin Tran, Dustin Zelle, Eric Noland, Erwin Huizenga, Eugene Kharitonov, Frederick Liu, Gagik Amirkhanyan, Glenn Cameron, Hadi Hashemi, Hanna Klimczak-Plucińska, Harman Singh, Harsh Mehta, Harshal~Tushar Lehri, Hussein Hazimeh, Ian Ballantyne, Idan Szpektor, Ivan Nardini, Jean Pouget-Abadie, Jetha Chan, Joe Stanton, John Wieting, Jonathan Lai, Jordi Orbay, Joseph Fernandez, Josh Newlan, Ju~yeong Ji, Jyotinder Singh, Kat Black, Kathy Yu, Kevin Hui, Kiran Vodrahalli, Klaus Greff, Linhai Qiu, Marcella Valentine, Marina Coelho, Marvin Ritter, Matt Hoffman, Matthew Watson, Mayank Chaturvedi, Michael Moynihan, Min Ma, Nabila Babar, Natasha Noy, Nathan Byrd, Nick Roy, Nikola Momchev, Nilay Chauhan, Noveen Sachdeva, Oskar
  Bunyan, Pankil Botarda, Paul Caron, Paul~Kishan Rubenstein, Phil Culliton, Philipp Schmid, Pier~Giuseppe Sessa, Pingmei Xu, Piotr Stanczyk, Pouya Tafti, Rakesh Shivanna, Renjie Wu, Renke Pan, Reza Rokni, Rob Willoughby, Rohith Vallu, Ryan Mullins, Sammy Jerome, Sara Smoot, Sertan Girgin, Shariq Iqbal, Shashir Reddy, Shruti Sheth, Siim Põder, Sijal Bhatnagar, Sindhu~Raghuram Panyam, Sivan Eiger, Susan Zhang, Tianqi Liu, Trevor Yacovone, Tyler Liechty, Uday Kalra, Utku Evci, Vedant Misra, Vincent Roseberry, Vlad Feinberg, Vlad Kolesnikov, Woohyun Han, Woosuk Kwon, Xi~Chen, Yinlam Chow, Yuvein Zhu, Zichuan Wei, Zoltan Egyed, Victor Cotruta, Minh Giang, Phoebe Kirk, Anand Rao, Kat Black, Nabila Babar, Jessica Lo, Erica Moreira, Luiz~Gustavo Martins, Omar Sanseviero, Lucas Gonzalez, Zach Gleicher, Tris Warkentin, Vahab Mirrokni, Evan Senter, Eli Collins, Joelle Barral, Zoubin Ghahramani, Raia Hadsell, Yossi Matias, D.~Sculley, Slav Petrov, Noah Fiedel, Noam Shazeer, Oriol Vinyals, Jeff Dean, Demis Hassabis,
  Koray Kavukcuoglu, Clement Farabet, Elena Buchatskaya, Jean-Baptiste Alayrac, Rohan Anil, Dmitry Lepikhin, Sebastian Borgeaud, Olivier Bachem, Armand Joulin, Alek Andreev, Cassidy Hardin, Robert Dadashi, and Léonard Hussenot.
\newblock Gemma 3 technical report.
\newblock arXiv preprint arXiv:2503.19786, 2025.

\bibitem{wei2022chain}
Jason Wei, Xuezhi Wang, Dale Schuurmans, Maarten Bosma, Brian Ichter, Fei Xia, Ed~H. Chi, Quoc~V. Le, and Denny Zhou.
\newblock Chain-of-thought prompting elicits reasoning in large language models.
\newblock In {\em Advances in Neural Information Processing Systems}, 2022.

\bibitem{xiao2024logicvista}
Yijia Xiao, Edward Sun, Tianyu Liu, and Wei Wang.
\newblock Logicvista: Multimodal llm logical reasoning benchmark in visual contexts.
\newblock {\em arXiv preprint arXiv:2407.04973}, 2024.

\bibitem{xu2024mcbench}
Yunqiu Xu, Linchao Zhu, and Yi~Yang.
\newblock Mc-bench: A benchmark for multi-context visual grounding in the era of mllms.
\newblock {\em arXiv preprint arXiv:2410.12332}, 2024.

\bibitem{yao2022react}
Shinn Yao, Jiachang Zhao, Dian Yu, Shuyang Gao, Yujie Chen, Zhou Yu, and Karthik Narasimhan.
\newblock React: Synergizing reasoning and acting in language models.
\newblock In {\em Advances in Neural Information Processing Systems}, 2022.

\bibitem{yu2023mmvet}
Weihao Yu, Zhengyuan Yang, Lingfeng Ren, Linjie Li, Jianfeng Wang, Kevin Lin, Chung-Ching Lin, Zicheng Liu, Lijuan Wang, and Xinchao Wang.
\newblock Mm-vet: Evaluating large multimodal models for integrated capabilities.
\newblock arXiv preprint arXiv:2308.02490, 2023.

\bibitem{mmmu}
Xiang Yue, Yuansheng Ni, Kai Zhang, Tianyu Zheng, Ruoqi Liu, Ge~Zhang, Samuel Stevens, Dongfu Jiang, Weiming Ren, Yuxuan Sun, Cong Wei, Botao Yu, Ruibin Yuan, Renliang Sun, Ming Yin, Boyuan Zheng, Zhenzhu Yang, Yibo Liu, Wenhao Huang, Huan Sun, Yu~Su, and Wenhu Chen.
\newblock Mmmu: A massive multi-discipline multimodal understanding and reasoning benchmark for expert agi.
\newblock \url{https://arxiv.org/abs/2311.16502}, 2023.
\newblock Accessed: 2025-05-14.

\bibitem{zellers2019vcr}
Rowan Zellers, Yonatan Bisk, Ali Farhadi, and Yejin Choi.
\newblock From recognition to cognition: Visual commonsense reasoning.
\newblock In {\em Proceedings of the IEEE/CVF Conference on Computer Vision and Pattern Recognition (CVPR)}, pages 6720--6730, 2019.

\end{thebibliography}

  \appendix
  \onecolumn
  
\newpage
\appendix
\onecolumn
\section{Basic Settings}

\begin{figure}[h]
\centering
\includegraphics[width=1\textwidth]{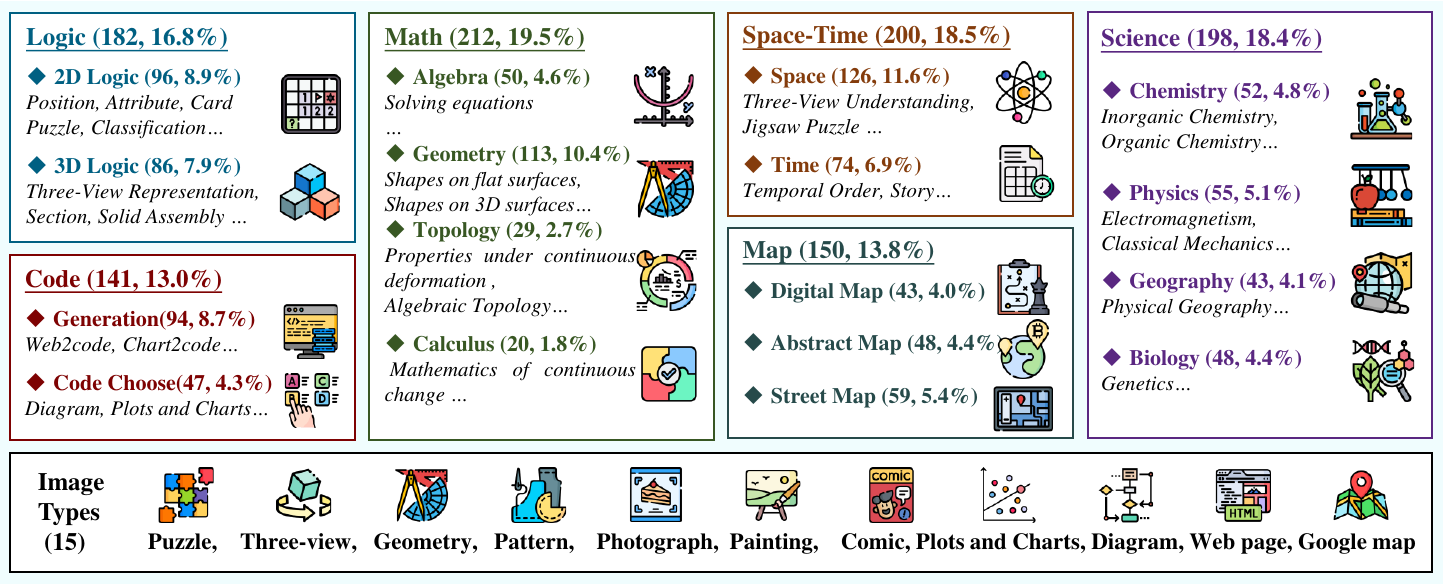}
\caption{Task-type distribution and sub-category breakdown across the \MMMR~benchmark.}
\label{fig:data_summary}
\end{figure}

Figure~\ref{fig:data_summary} provides a comprehensive summary of the \MMMR~benchmark, including task type distributions, instance counts, and sub-category breakdowns. The benchmark consists of 1,083 multi-modal reasoning tasks spanning six high-level domains: Logic, Math, Space-Time, Code, Map, and Science. Each domain contains diverse subtypes (e.g., deductive logic, algebraic manipulation, spatial tracking, etc.), designed to probe distinct reasoning faculties. The chart further reports the proportion of each task type, ensuring balanced but realistic coverage aligned with real-world reasoning demands.

\textbf{Expert Baselines.}\label{expert}~
To contextualize the capabilities of MLLMs and MLLMs-T, we introduce two upper-bound baselines referred to as \emph{Experts}, representing different degrees of human involvement:

\begin{itemize}
  \item \textbf{Expert (Human only)}: This setting represents pure human reasoning without any model assistance. We selected three co-authors of this paper, each with graduate-level expertise in AI, cognitive science, or related fields, to independently solve the benchmark tasks. Participants were provided with full task descriptions and multi-modal inputs (text and images), but were not exposed to model outputs or allowed external tools. To ensure reliability, each sample was independently answered by at least two annotators; disagreements were resolved via majority voting. The inter-annotator agreement, measured by Krippendorff’s alpha, reached 0.84, indicating high consistency and shared task understanding.

  \item \textbf{Expert (Human + GPT-4o)}: This hybrid configuration simulates a human-in-the-loop decision-support paradigm, where the same human experts were allowed to optionally consult GPT-4o during task resolution. Annotators first formed an independent judgment, then optionally queried GPT-4o for additional insights or solutions. Final responses reflected either acceptance or revision of GPT-4o’s suggestions, along with justifications. This setting measures the upper bound of human-AI collaboration in structured reasoning tasks.
\end{itemize}

These expert configurations serve as practical performance ceilings: the human-only setting captures unaided expert cognition, while the hybrid setting reflects augmented performance with access to state-of-the-art MLLM support. Together, they frame the evaluation of MLLMs within a broader continuum of human-machine reasoning capabilities.

\newpage

\section{Prompt Design}\label{prompts}

\subsection{Base}\label{app:pmt_base}

\noindent\textbf{Prompt Example.}~~Base’s prompt example is as follows:

\begin{tcolorbox}[
  colback=gray!10,      
  colframe=black,  
  arc=1mm,
  boxrule=0.5mm,
  left=6pt,
  right=6pt,
  top=6pt,
  bottom=6pt,
  title=\textbf{Base’s prompt example},
  before skip=6pt,
  after skip=6pt,
  breakable
]
\textbf{Zero-shot Prompt:} \\
$\blacktriangleright$\textbf{\{question\}}\\
$\blacktriangleright$\textbf{\{choice\}}\\
Please provide the final answer and store it in \textcolor{red}{\textbf{\texttt{\textbackslash boxed\{answer\}}}}. \\
\textbf{Critique Prompt:} Review your previous answer and find problems with your answer. \\
\textbf{Improve Prompt:} Based on the problems you found, improve your answer. Please reiterate your answer, with your final answer a single numerical number, In the form \textcolor{red}{\textbf{\texttt{\textbackslash boxed\{answer\}}}}.
\end{tcolorbox}

\subsection{Thinking Prompt}\label{app:pmt_thinking}
\noindent\textbf{Prompt Example.}~~The Thinking prompt is as follows:
\begin{tcolorbox}[
  colback=gray!10,      
  colframe=black,  
  arc=1mm,
  boxrule=0.5mm,
  left=6pt,
  right=6pt,
  top=6pt,
  bottom=6pt,
  title=\textbf{Thinking Prompt Example},
  before skip=6pt,
  after skip=6pt,
  breakable
]
$\blacktriangleright$\textbf{\{question\}}\\
$\blacktriangleright$\textbf{\{choice\}}\\
Please think deeply before your response.\\
Please provide the final answer and store it in \textcolor{red}{\textbf{\texttt{\textbackslash boxed\{answer\}}}}.
\end{tcolorbox}

\subsection{Image-text to Text Prompt}\label{app:pmt_img2text}
\noindent\textbf{Prompt Example.}~~The Image-text to text prompt is as follows:
\begin{tcolorbox}[
  colback=gray!10,      
  colframe=black,  
  arc=1mm,
  boxrule=0.5mm,
  left=6pt,
  right=6pt,
  top=6pt,
  bottom=6pt,
  title=\textbf{Image-text to Text Prompt Example},
  before skip=6pt,
  after skip=6pt,
  breakable
]
$\blacktriangleright$Based on the question and the image, please summary it in pure text. Just summary the question and image as detailed as possible, no need to give the answer.
\end{tcolorbox}

\subsection{Random Choice Baseline}\label{app:pmt_random_choice}
\noindent\textbf{Implementation Logic.}~~The logic for the Random Choice baseline is as follows:
\begin{tcolorbox}[
  colback=gray!10,
  colframe=black,
  arc=1mm,
  boxrule=0.5mm,
  left=6pt,
  right=6pt,
  top=6pt,
  bottom=6pt,
  title=\textbf{Random Choice Baseline Logic},
  before skip=6pt,
  after skip=6pt,
  breakable,
  before skip=6pt,
  after skip=6pt,
  breakable,
  listing engine=listings, 
  listing options={basicstyle=\ttfamily\footnotesize, breaklines=true} 
]
def random\_choice\_baseline(questions, output\_file):\\
\hspace*{1em}with open(output\_file, 'w', encoding='utf-8') as f\_out:\\
\hspace*{2em}for q in questions:\\
\hspace*{3em}n = len(q["choices"])\\
\hspace*{3em}labels = [chr(ord('A') + i) for i in range(n)]\\
\hspace*{3em}prediction = random.choice(labels)\\
\hspace*{3em}correct = normalize\_answer(q["correct"])\\
\\
\hspace*{3em}result = \{\\
\hspace*{4em}"question": q.get("question", ""),\\
\hspace*{4em}"prediction": prediction,\\
\hspace*{4em}"correct": correct\\
\hspace*{3em}\}\\
\hspace*{3em}f\_out.write(json.dumps(result, ensure\_ascii=False))
\end{tcolorbox}

\subsection{Frequent Choice Baseline}\label{app:pmt_frequent_choice}
\noindent\textbf{Implementation Logic.}~~The logic for the Frequent Choice baseline is as follows:
\begin{tcolorbox}[
  colback=gray!10,      
  colframe=black,  
  arc=1mm,
  boxrule=0.5mm,
  left=6pt,
  right=6pt,
  top=6pt,
  bottom=6pt,
  title=\textbf{Frequent Choice Baseline Logic},
  before skip=6pt,
  after skip=6pt,
  breakable,
  listing engine=listings, 
  listing options={basicstyle=\ttfamily\footnotesize, breaklines=true} 
]
def frequent\_choice\_baseline(questions, output\_file):\\
\hspace*{1em}counter = Counter()\\
\hspace*{1em}for q in questions:\\
\hspace*{2em}correct = normalize\_answer(q["correct"])\\
\hspace*{2em}if correct:\\
\hspace*{3em}counter[correct] += 1\\
\\
\hspace*{1em}if not counter:\\
\hspace*{2em}print("No valid answers found for frequent choice baseline. Skip.")\\
\hspace*{2em}return\\
\\
\hspace*{1em}most\_common\_choice, \_ = counter.most\_common(1)[0]\\
\\
\hspace*{1em}with open(output\_file, 'w', encoding='utf-8') as f\_out:\\
\hspace*{2em}for q in questions:\\
\hspace*{3em}correct = normalize\_answer(q["correct"])\\
\hspace*{3em}result = \{\\
\hspace*{4em}"question": q.get("question", ""),\\
\hspace*{4em}"prediction": most\_common\_choice,\\
\hspace*{4em}"correct": correct\\
\hspace*{3em}\}\\
\hspace*{3em}f\_out.write(json.dumps(result, ensure\_ascii=False))
\end{tcolorbox}

\newpage

\section{Case Study}

\makebox[0.5em][l]{} 

\textcolor{red}{1. Logic: Thinking Case} 
\dotfill 
\textcolor{red}{\pageref{fig:Logic-Thinking}}

\textcolor{red}{2D Logic: Non-Thinking Case} 
\dotfill 
\textcolor{red}{\pageref{fig:2D Logic}}

\textcolor{red}{3D Logic: Non-Thinking Case} 
\dotfill 
\textcolor{red}{\pageref{fig:3D Logic}}

\textcolor{red}{2. Code: Thinking Case} 
\dotfill 
\textcolor{red}{\pageref{fig:Code-Thinking}}

\textcolor{red}{Generation: Non-Thinking Case} 
\dotfill 
\textcolor{red}{\pageref{fig:Generation}}

\textcolor{red}{Code Choose: Non-Thinking Case} 
\dotfill 
\textcolor{red}{\pageref{fig:Code Choose}}

\textcolor{red}{3. Math: Thinking Case} 
\dotfill 
\textcolor{red}{\pageref{fig:Math-Thinking}}

\textcolor{red}{Algebra: Non-Thinking Case} 
\dotfill 
\textcolor{red}{\pageref{fig:Algebra}}

\textcolor{red}{Geometry: Non-Thinking Case} 
\dotfill 
\textcolor{red}{\pageref{fig:Geometry}}

\textcolor{red}{Topology: Non-Thinking Case} 
\dotfill 
\textcolor{red}{\pageref{fig:TopoLogy}}

\textcolor{red}{Calculus: Non-Thinking Case} 
\dotfill 
\textcolor{red}{\pageref{fig:Calculus}}

\textcolor{red}{4. Space-Time: Thinking Case} 
\dotfill 
\textcolor{red}{\pageref{fig:Space-Time-Thinking}}

\textcolor{red}{Space Reasoning: Non-Thinking Case} 
\dotfill 
\textcolor{red}{\pageref{fig:Space Reasoning}}

\textcolor{red}{Time: Non-Thinking Case} 
\dotfill 
\textcolor{red}{\pageref{fig:Time}}

\textcolor{red}{5. Map: Thinking Case} 
\dotfill 
\textcolor{red}{\pageref{fig:Map-Thinking}}

\textcolor{red}{Route Plan: Non-Thinking Case} 
\dotfill 
\textcolor{red}{\pageref{fig:Route Plan}}

\textcolor{red}{Street Map: Non-Thinking Case} 
\dotfill 
\textcolor{red}{\pageref{fig:Street Map}}

\textcolor{red}{6. Science: Thinking Case} 
\dotfill 
\textcolor{red}{\pageref{fig:Science-Thinking}}

\textcolor{red}{Chemistry: Non-Thinking Case} 
\dotfill 
\textcolor{red}{\pageref{fig:Chemistry}}

\textcolor{red}{Physics: Non-Thinking Case} 
\dotfill 
\textcolor{red}{\pageref{fig:Physics}}

\textcolor{red}{Geography: Non-Thinking Case} 
\dotfill 
\textcolor{red}{\pageref{fig:Geography}}

\textcolor{red}{Biology: Non-Thinking Case} 
\dotfill 
\textcolor{red}{\pageref{fig:Biology}}

\begin{figure*}[hp]
    \centering
    \includegraphics[width=1\linewidth]{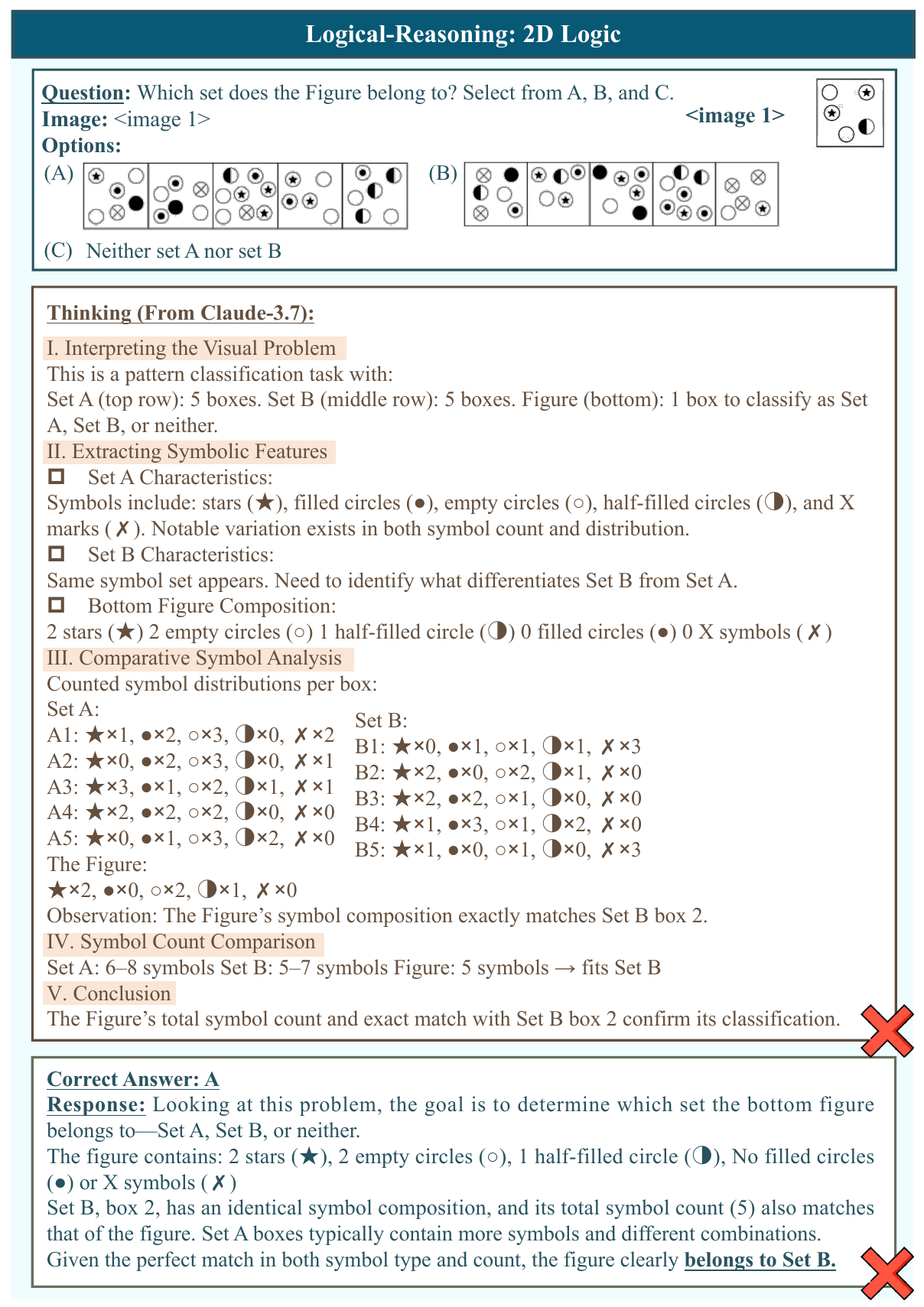}
    \vspace{-5mm}
    \caption{Logic: Thinking Case}
    \label{fig:Logic-Thinking}
\end{figure*}

\newpage
\begin{figure*}[hp]
    \centering
    \includegraphics[width=1\linewidth]{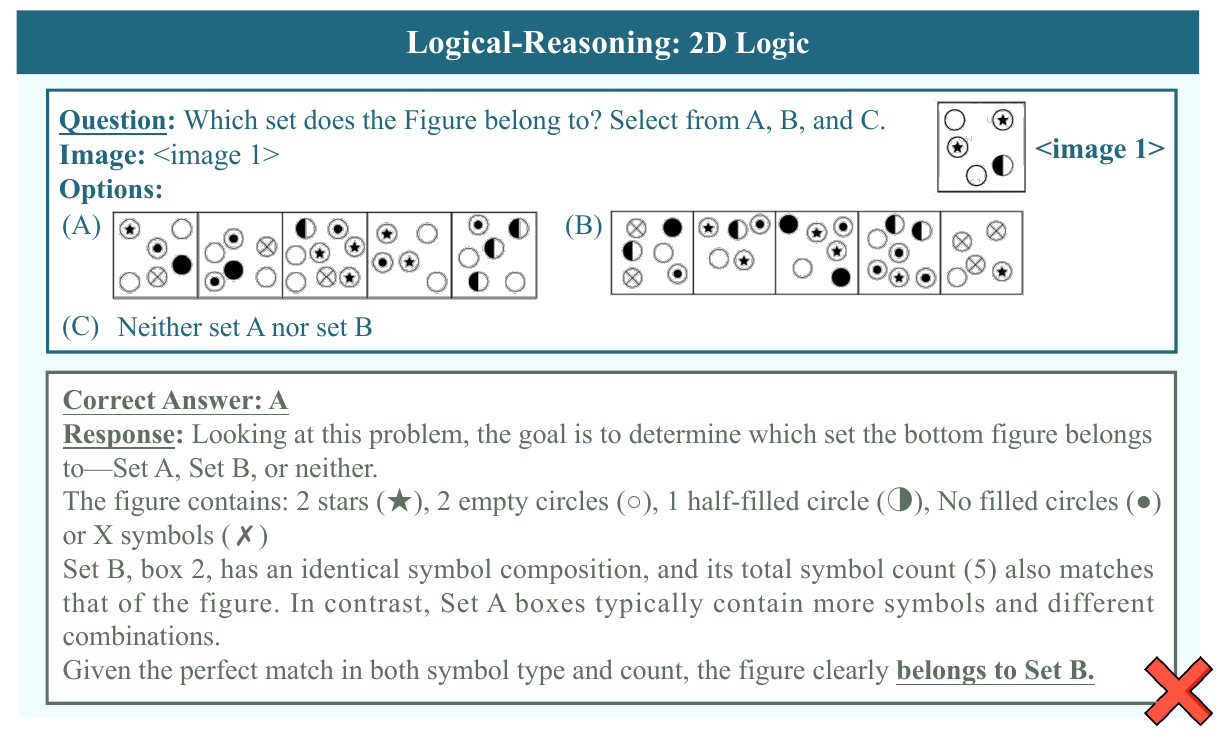}
    \vspace{-5mm}
    \caption{2D Logic: Non-Thinking Case}
    \label{fig:2D Logic}
\end{figure*}

\newpage
\begin{figure*}[hp]
    \centering
    \includegraphics[width=1\linewidth]{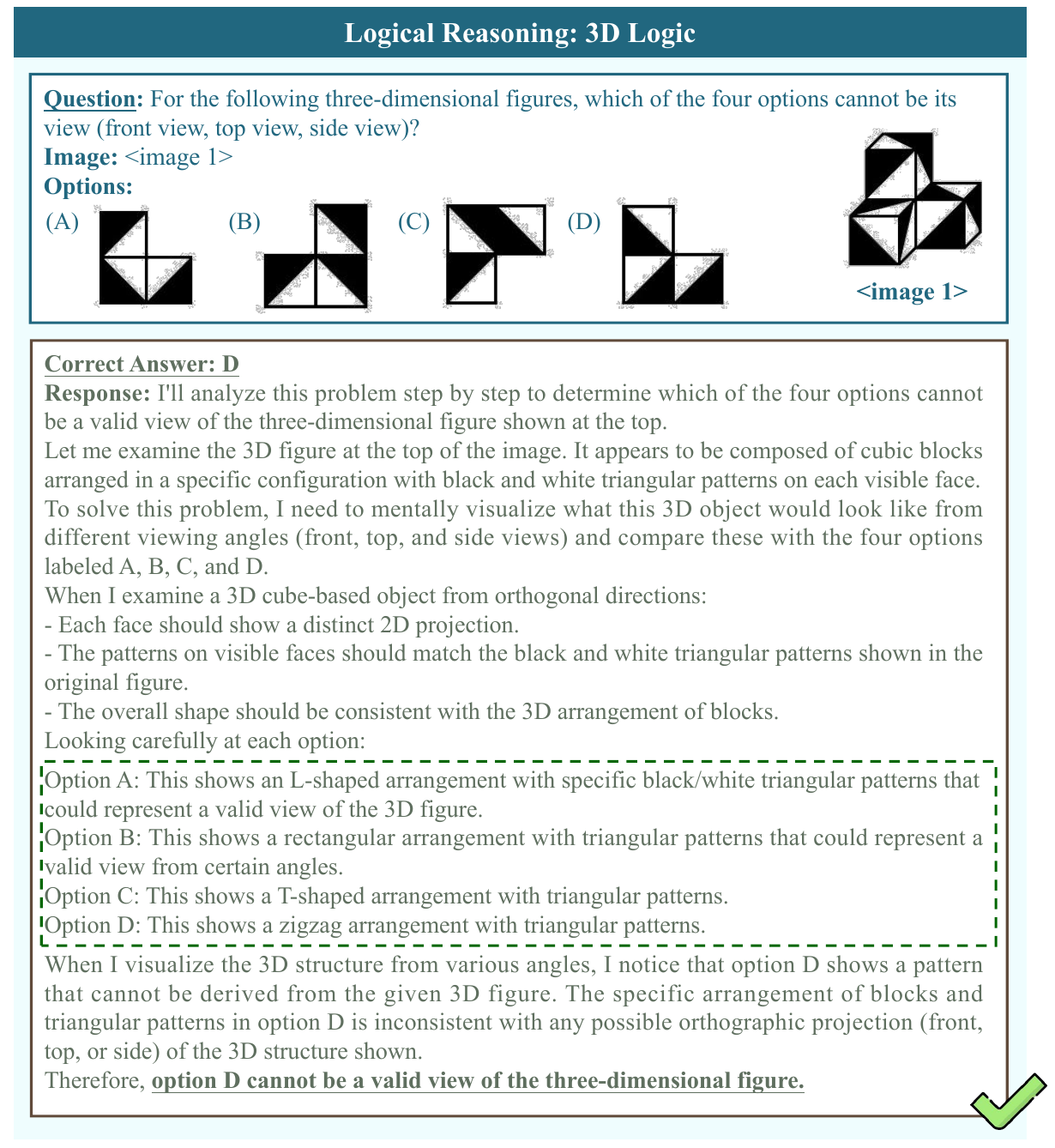}
    \vspace{-5mm}
    \caption{3D Logic: Non-Thinking Case}
    \label{fig:3D Logic}
\end{figure*}

\newpage
\begin{figure*}[hp]
    \centering
    \includegraphics[width=1\linewidth]{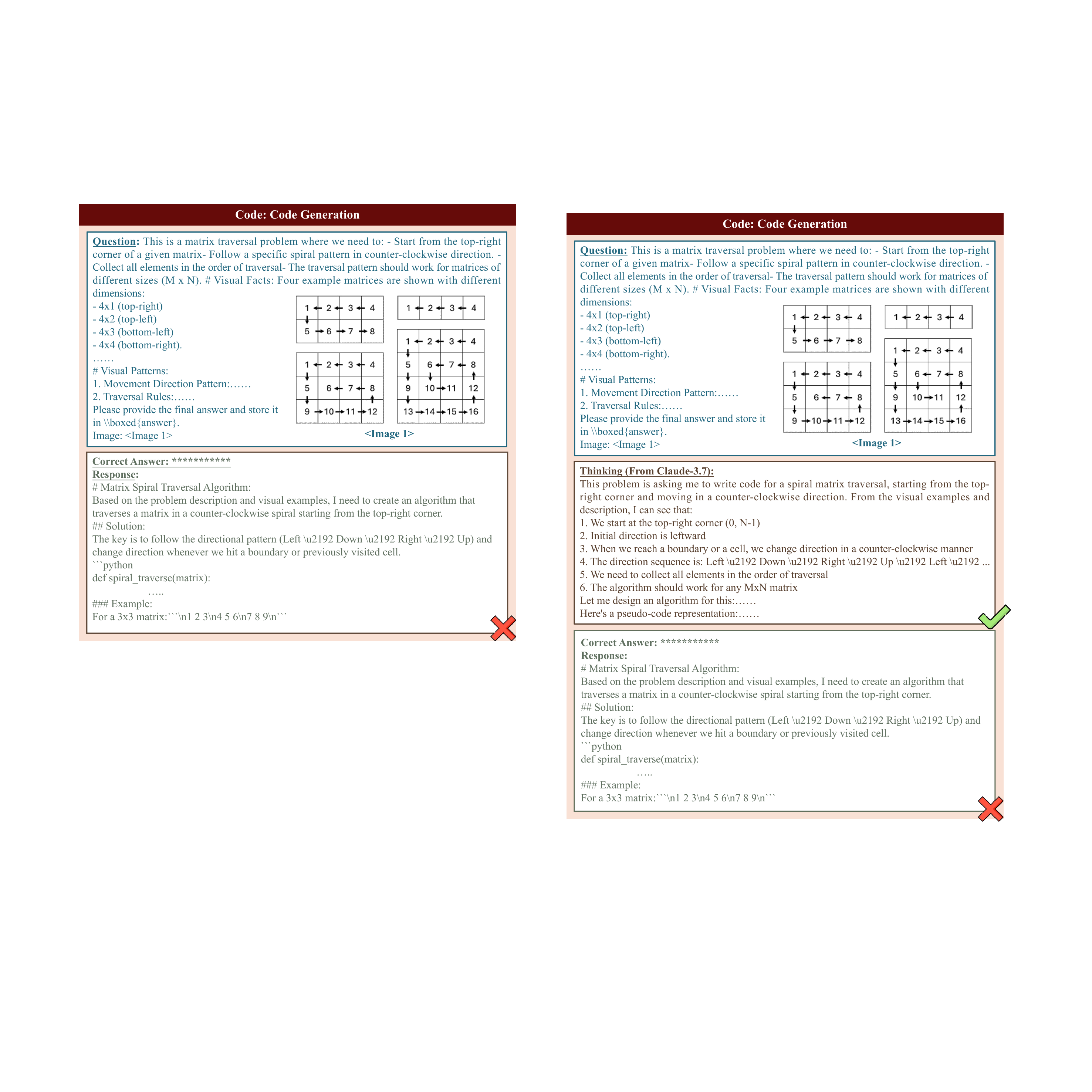}
    \vspace{-5mm}
    \caption{Code: Thinking Case}
    \label{fig:Code-Thinking}
\end{figure*}

\newpage
\begin{figure*}[hp]
    \centering
    \includegraphics[width=1\linewidth]{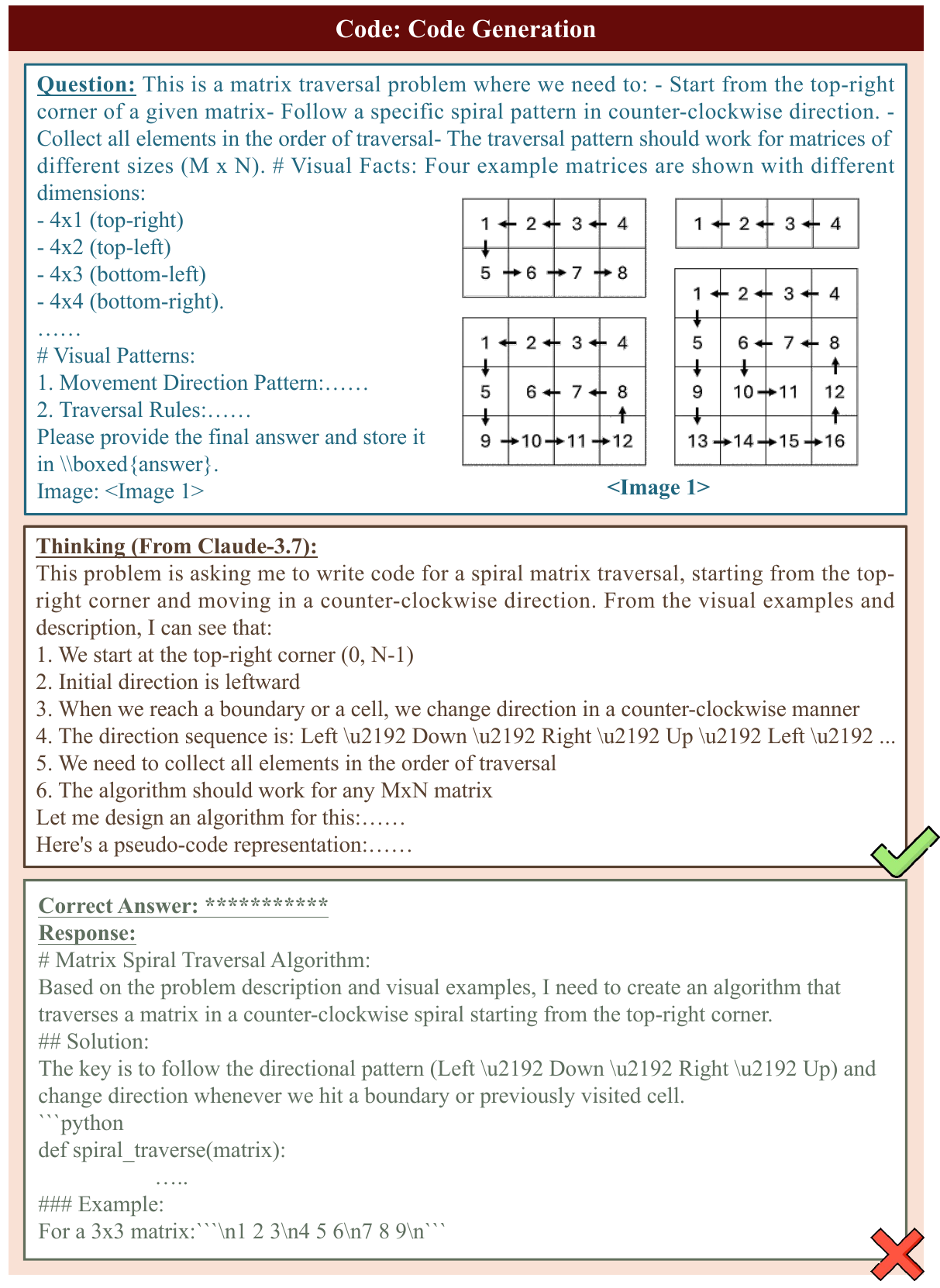}
    \vspace{-5mm}
    \caption{Generation: Non-Thinking Case}
    \label{fig:Generation}
\end{figure*}

\newpage
\begin{figure*}[hp]
    \centering
    \includegraphics[width=1\linewidth]{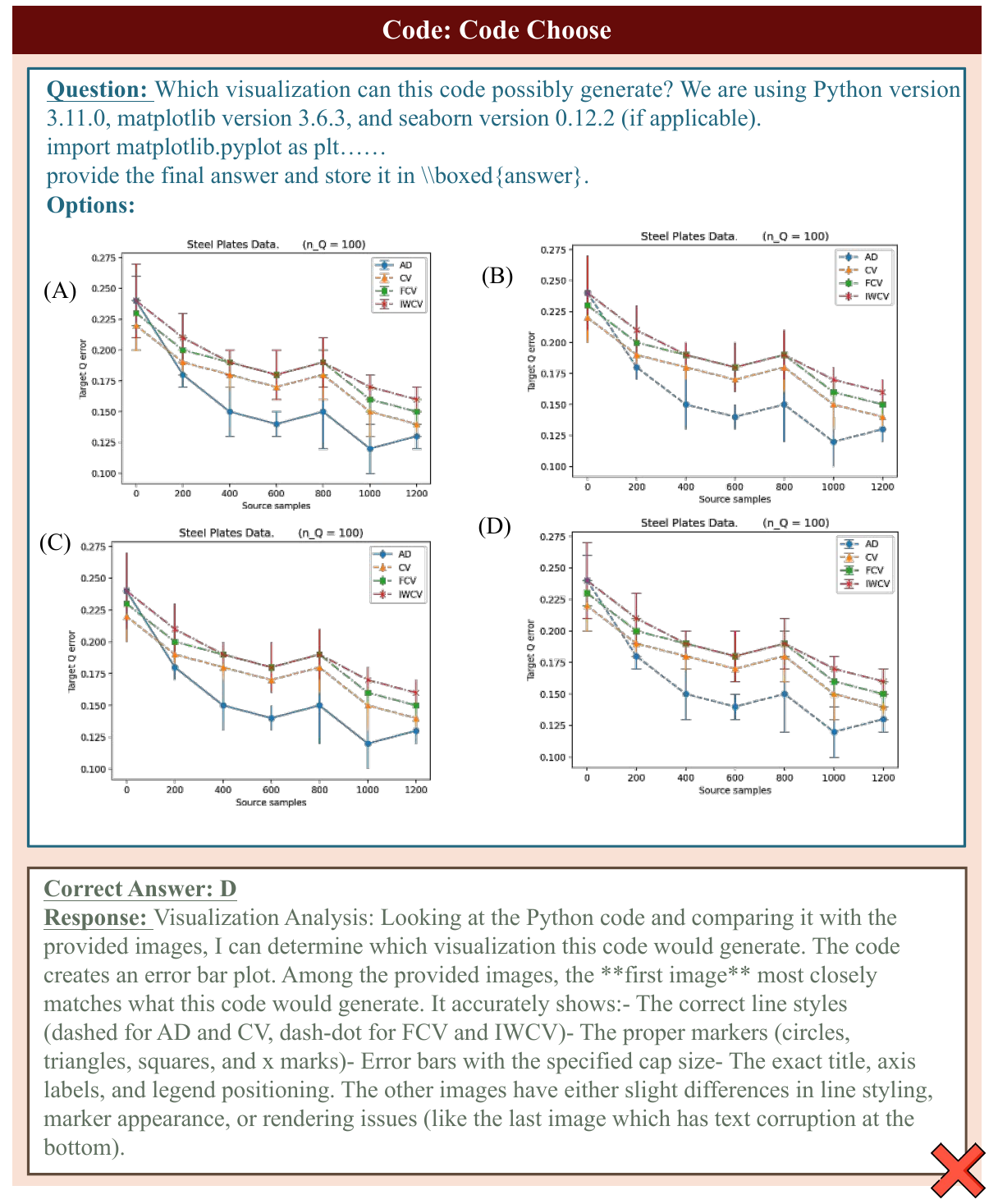}
    \vspace{-5mm}
    \caption{Code Choose: Non-Thinking Case}
    \label{fig:Code Choose}
\end{figure*}

\newpage
\begin{figure*}[hp]
    \centering
    \includegraphics[width=1\linewidth]{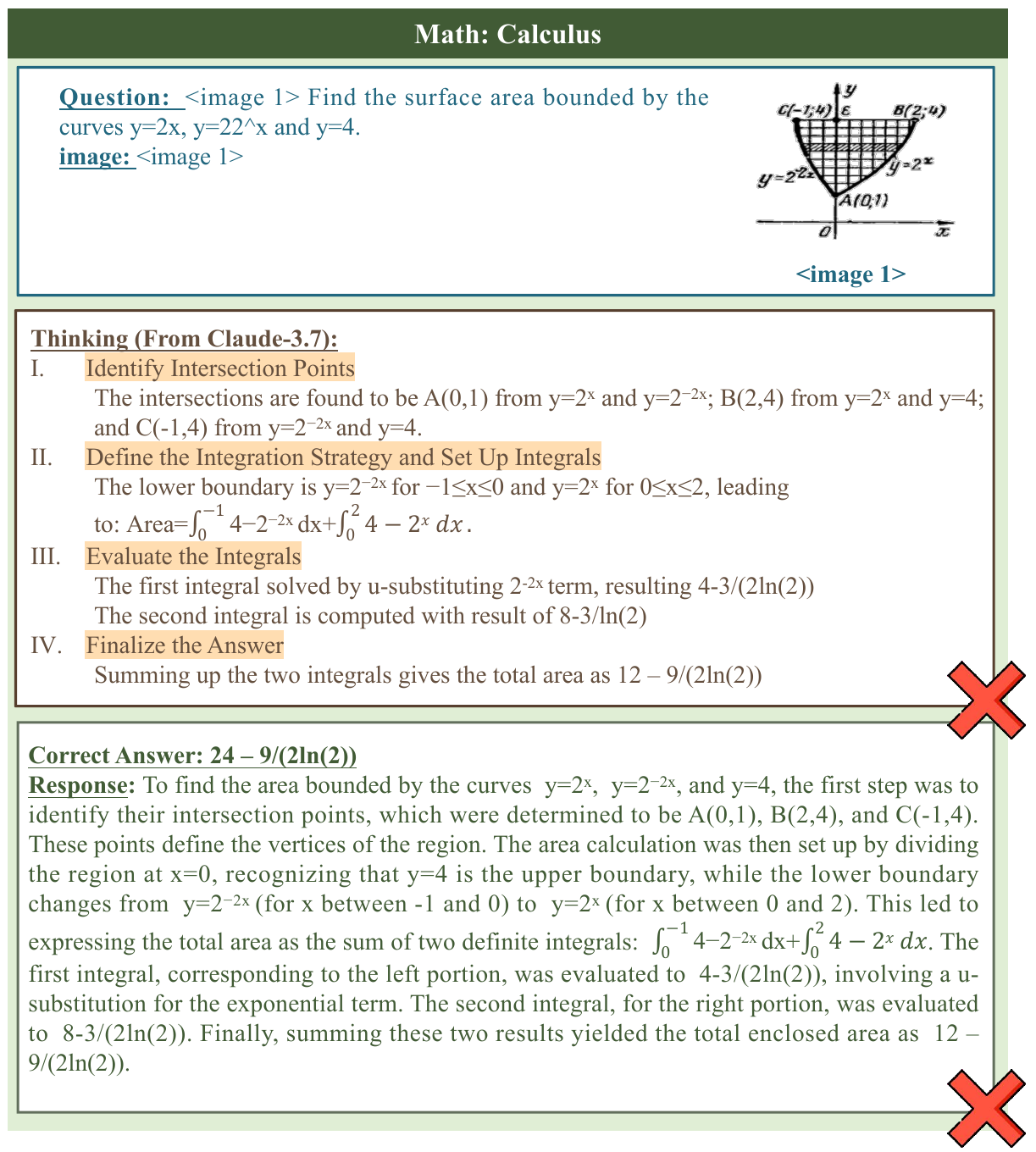}
    \vspace{-5mm}
    \caption{Math: Thinking Case}
    \label{fig:Math-Thinking}
\end{figure*}

\newpage
\begin{figure*}[hp]
    \centering
    \includegraphics[width=1\linewidth]{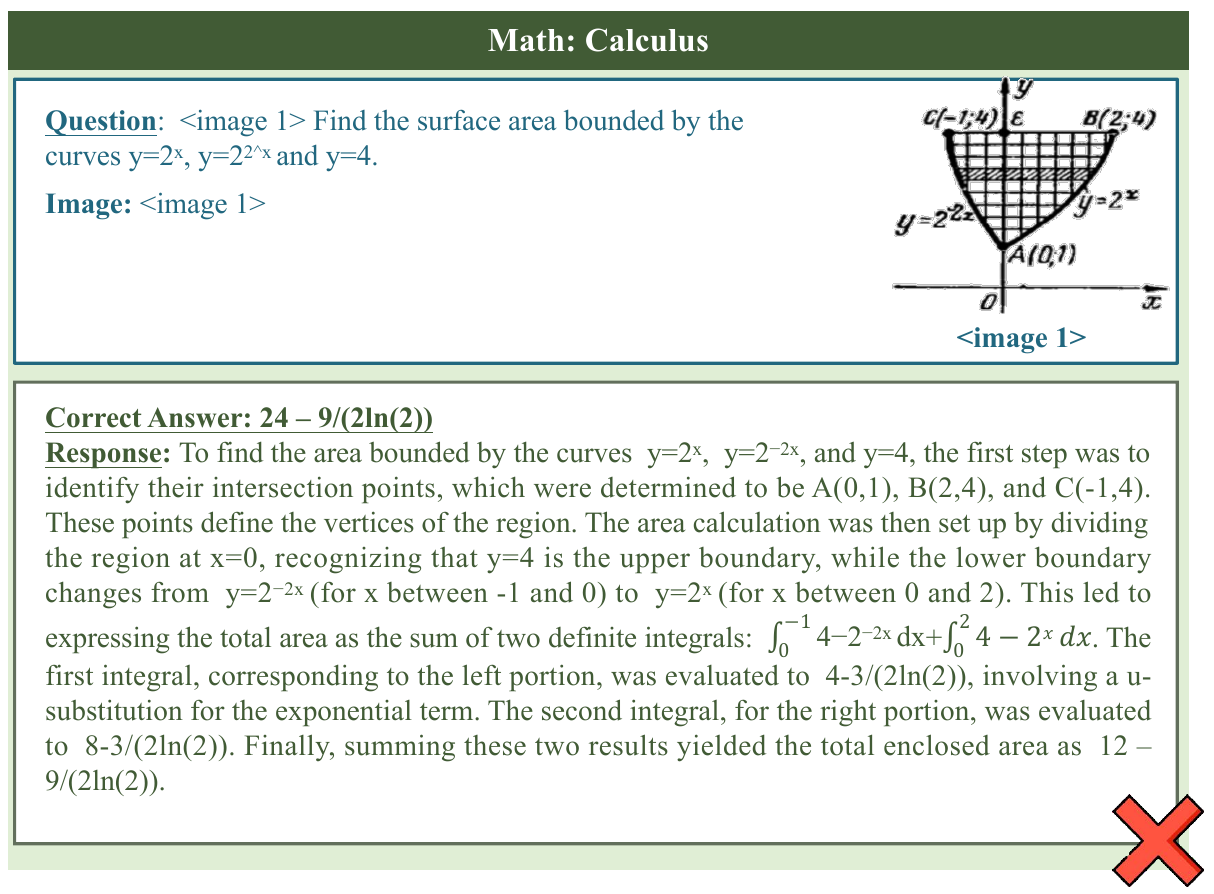}
    \vspace{-5mm}
    \caption{Algebra: Non-Thinking Case}
    \label{fig:Algebra}
\end{figure*}

\newpage
\begin{figure*}[hp]
    \centering
    \includegraphics[width=1\linewidth]{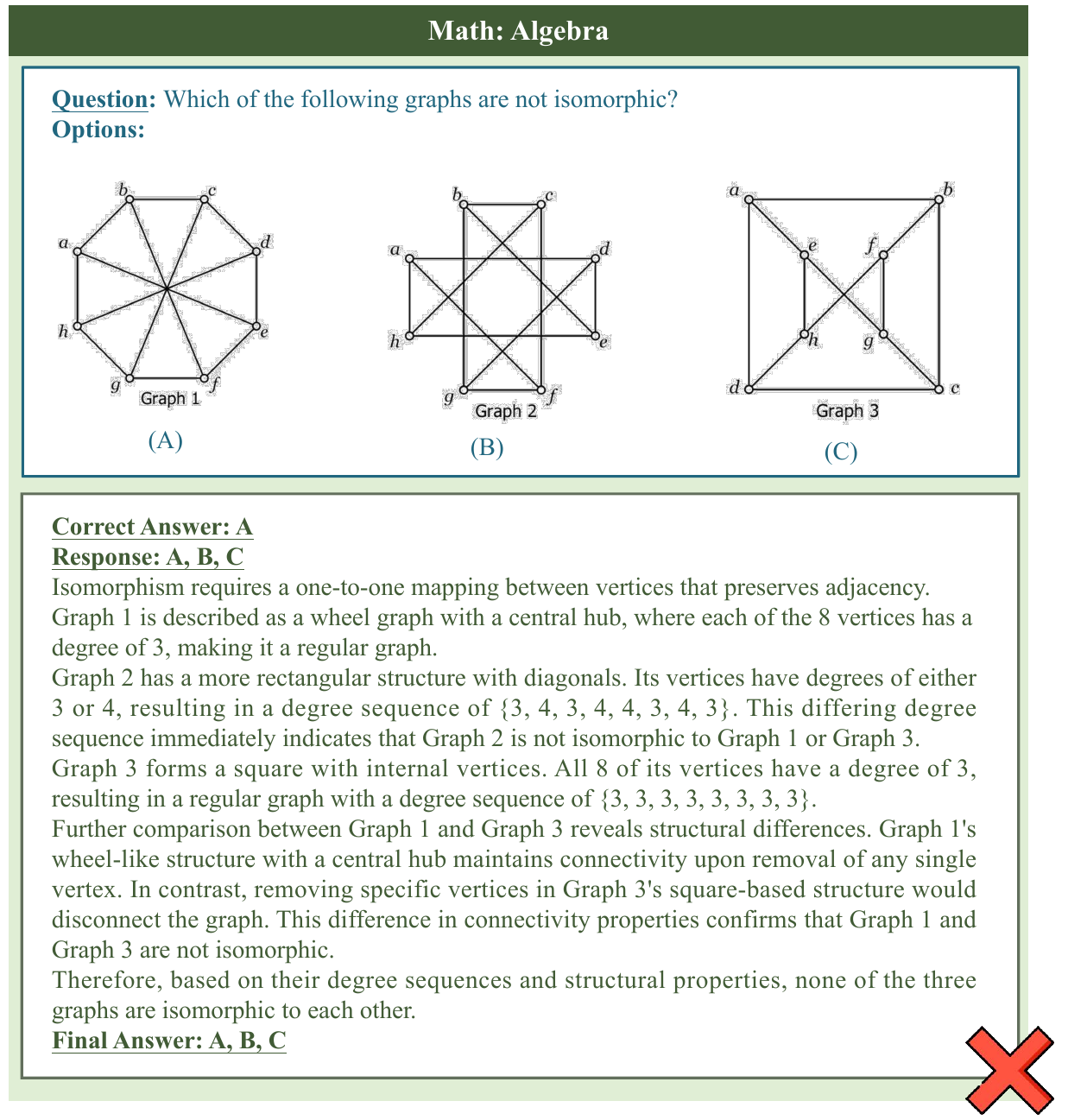}
    \vspace{-5mm}
    \caption{Geometry: Non-Thinking Case}
    \label{fig:Geometry}
\end{figure*}

\newpage
\begin{figure*}[hp]
    \centering
    \includegraphics[width=1\linewidth]{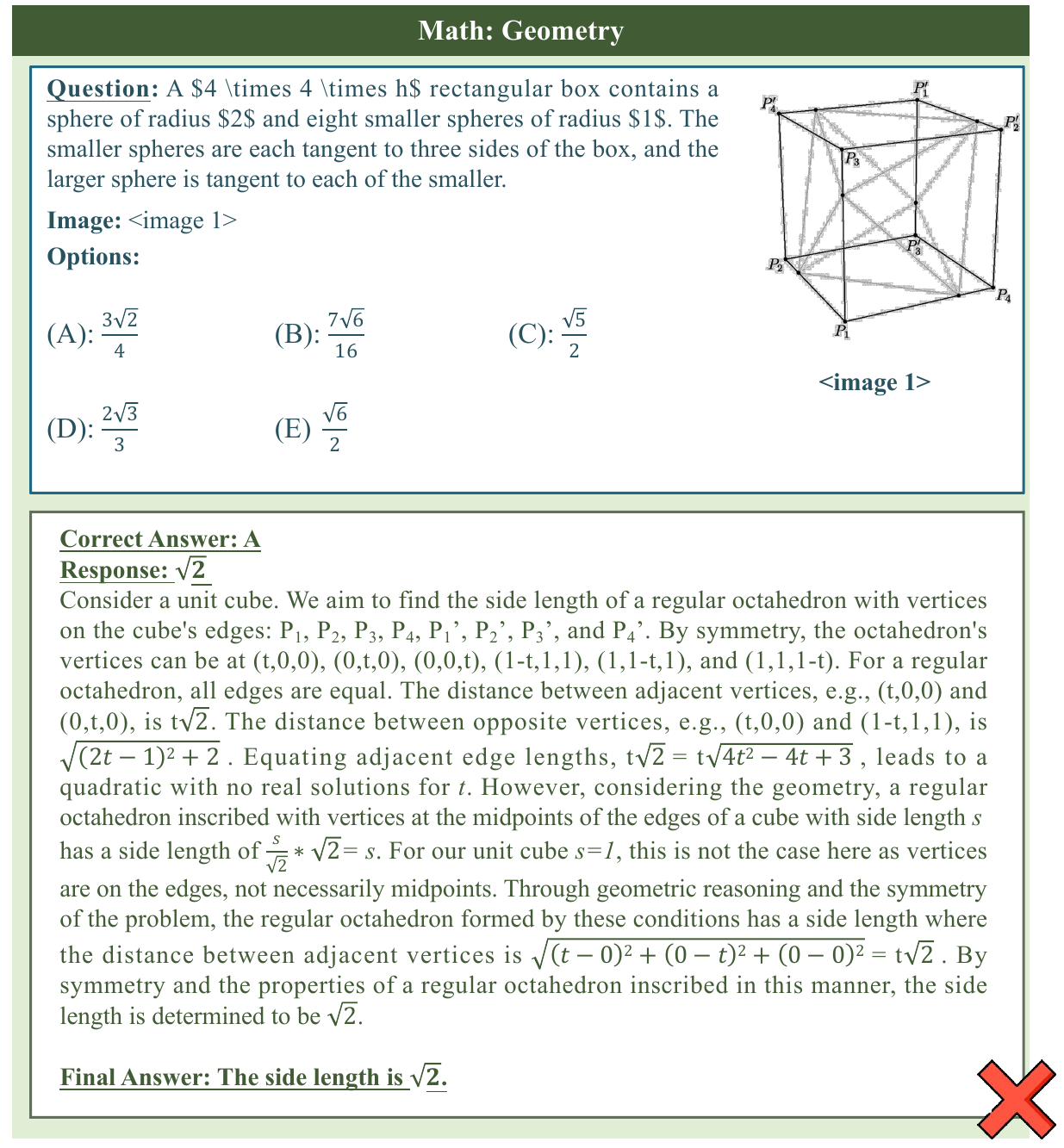}
    \vspace{-5mm}
    \caption{TopoLogy: Non-Thinking Case}
    \label{fig:TopoLogy}
\end{figure*}

\newpage
\begin{figure*}[hp]
    \centering
    \includegraphics[width=1\linewidth]{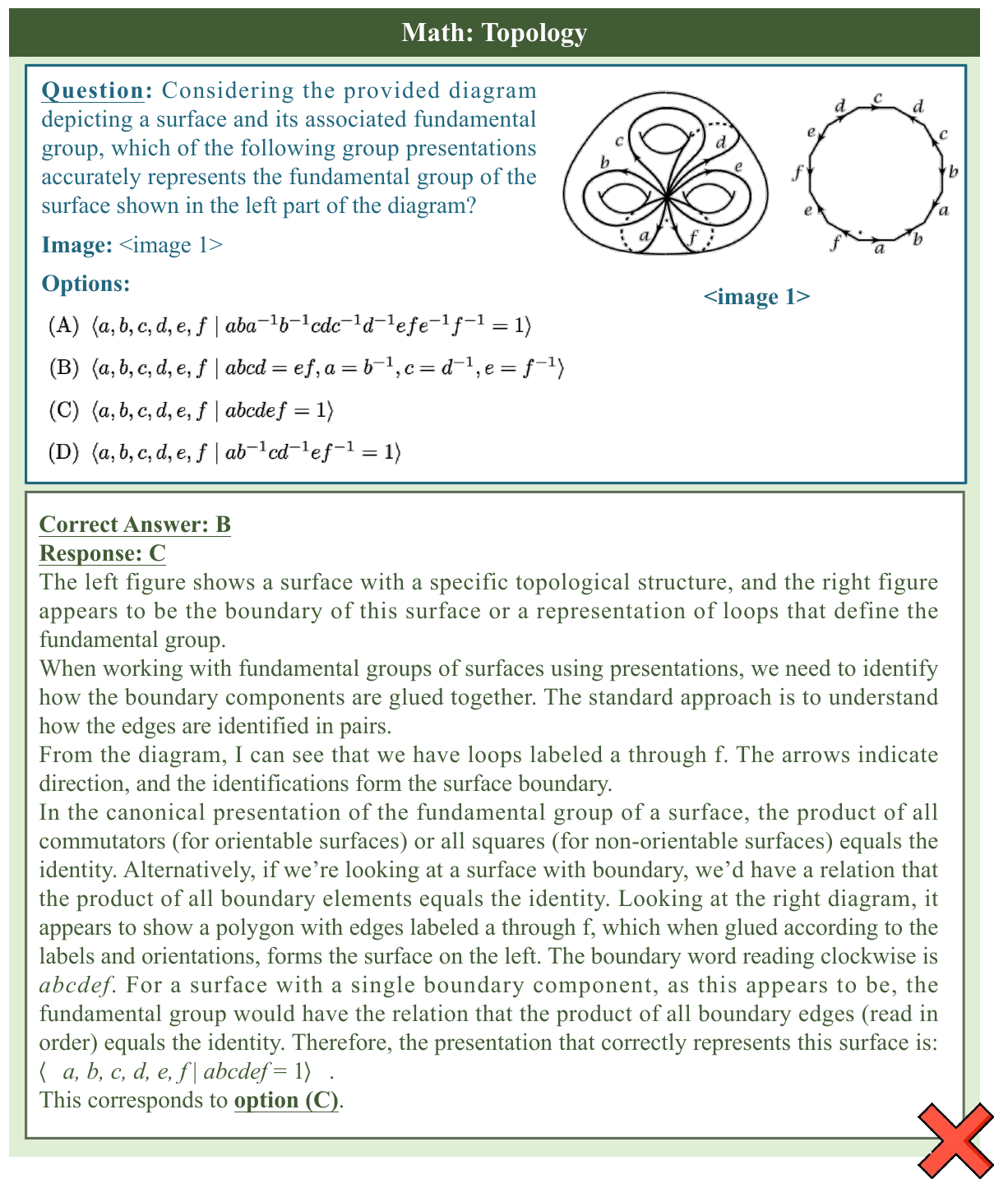}
    \vspace{-5mm}
    \caption{Calculus: Non-Thinking Case}
    \label{fig:Calculus}
\end{figure*}

\newpage
\begin{figure*}[hp]
    \centering
    \includegraphics[width=1\linewidth]{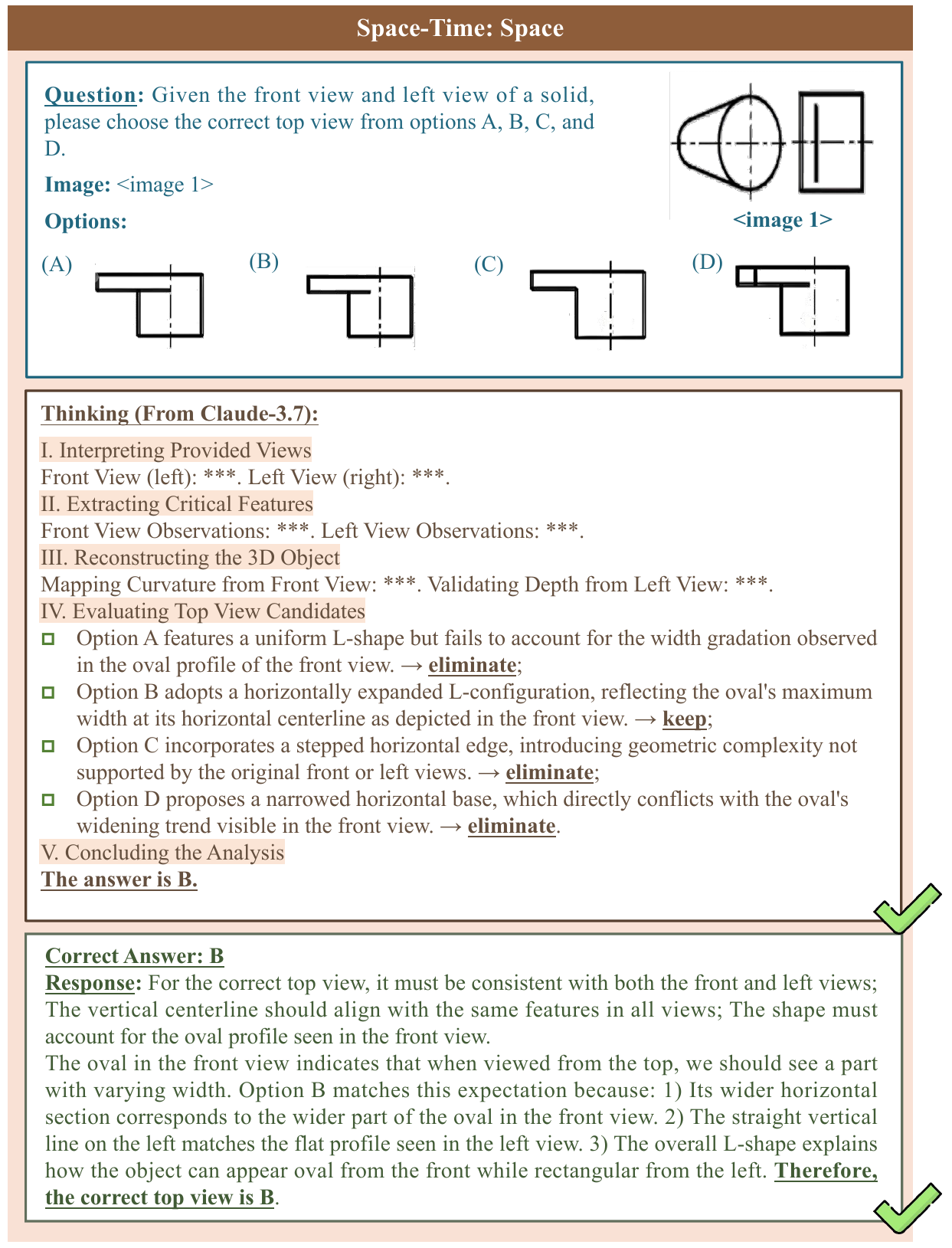}
    \vspace{-5mm}
    \caption{Space-Time: Thinking Case}
    \label{fig:Space-Time-Thinking}
\end{figure*}

\newpage
\begin{figure*}[hp]
    \centering
    \includegraphics[width=1\linewidth]{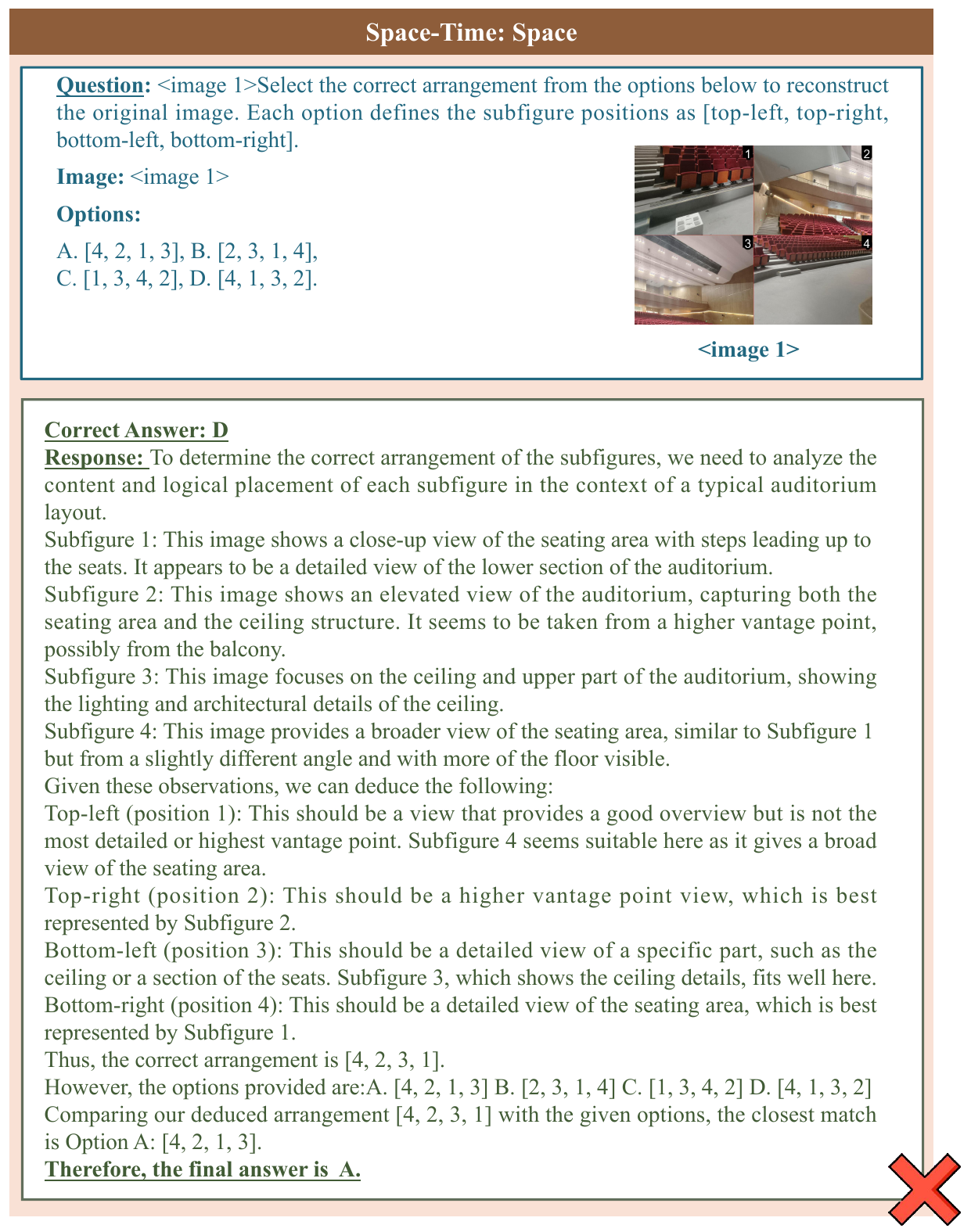}
    \vspace{-5mm}
    \caption{Space Reasoning: Non-Thinking Case}
    \label{fig:Space Reasoning}
\end{figure*}

\newpage
\begin{figure*}[hp]
    \centering
    \includegraphics[width=1\linewidth]{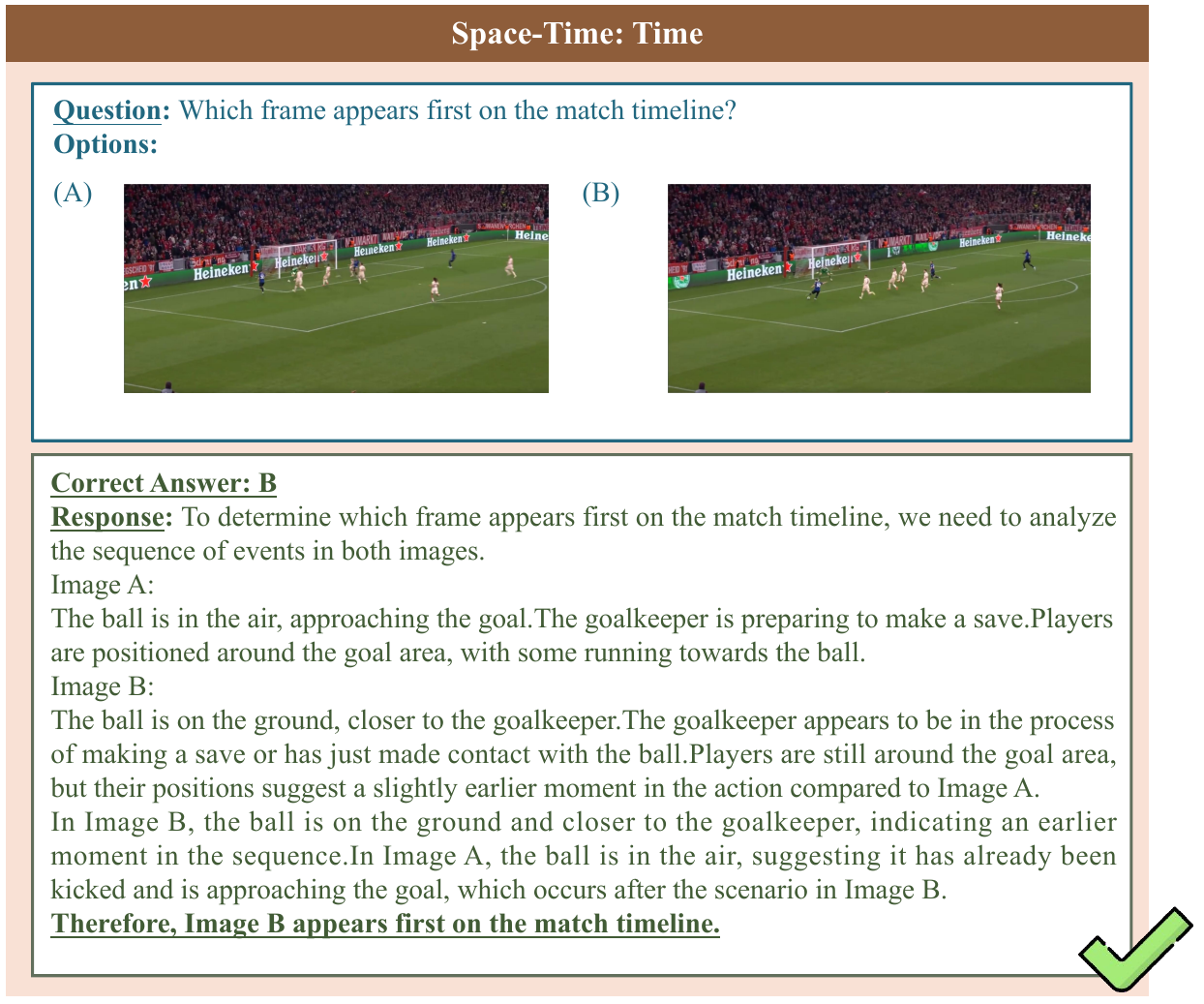}
    \vspace{-5mm}
    \caption{Time: Non-Thinking Case}
    \label{fig:Time}
\end{figure*}

\newpage
\begin{figure*}[hp]
    \centering
    \includegraphics[width=1\linewidth]{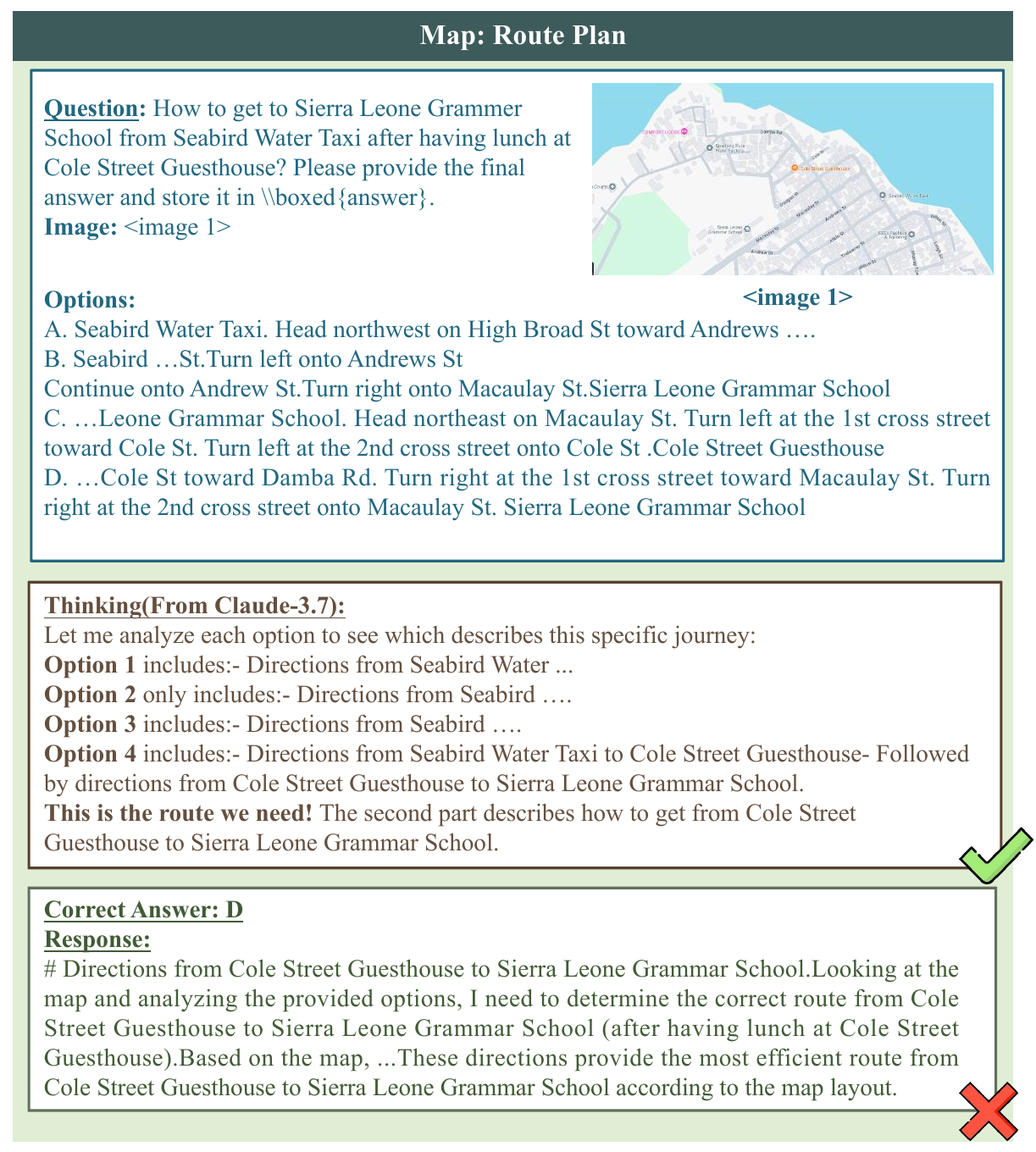}
    \vspace{-5mm}
    \caption{Map: Thinking Case}
    \label{fig:Map-Thinking}
\end{figure*}

\newpage
\begin{figure*}[hp]
    \centering
    \includegraphics[width=1\linewidth]{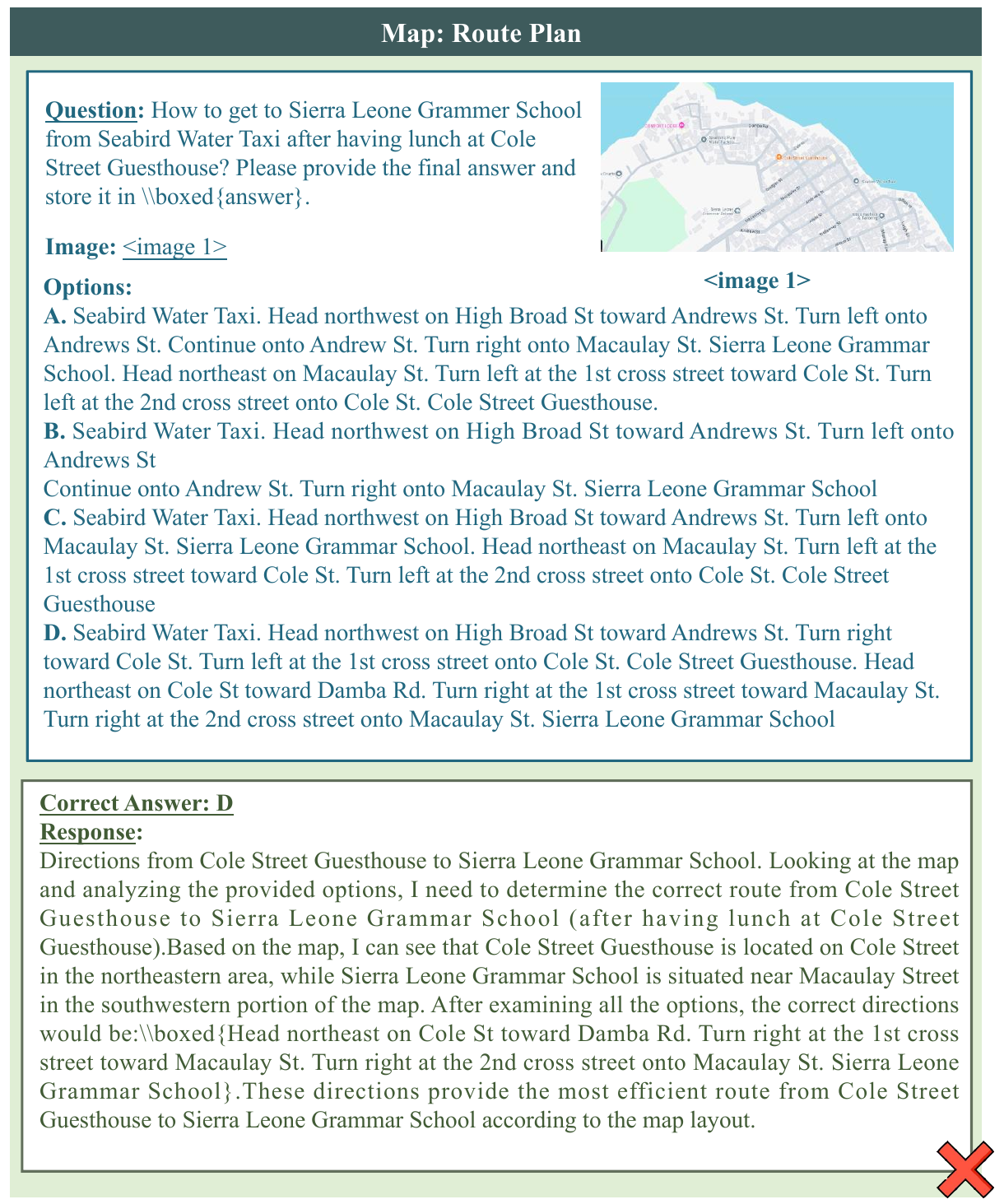}
    \vspace{-5mm}
    \caption{Route Plan: Non-Thinking Case}
    \label{fig:Route Plan}
\end{figure*}

\newpage
\begin{figure*}[hp]
    \centering
    \includegraphics[width=1\linewidth]{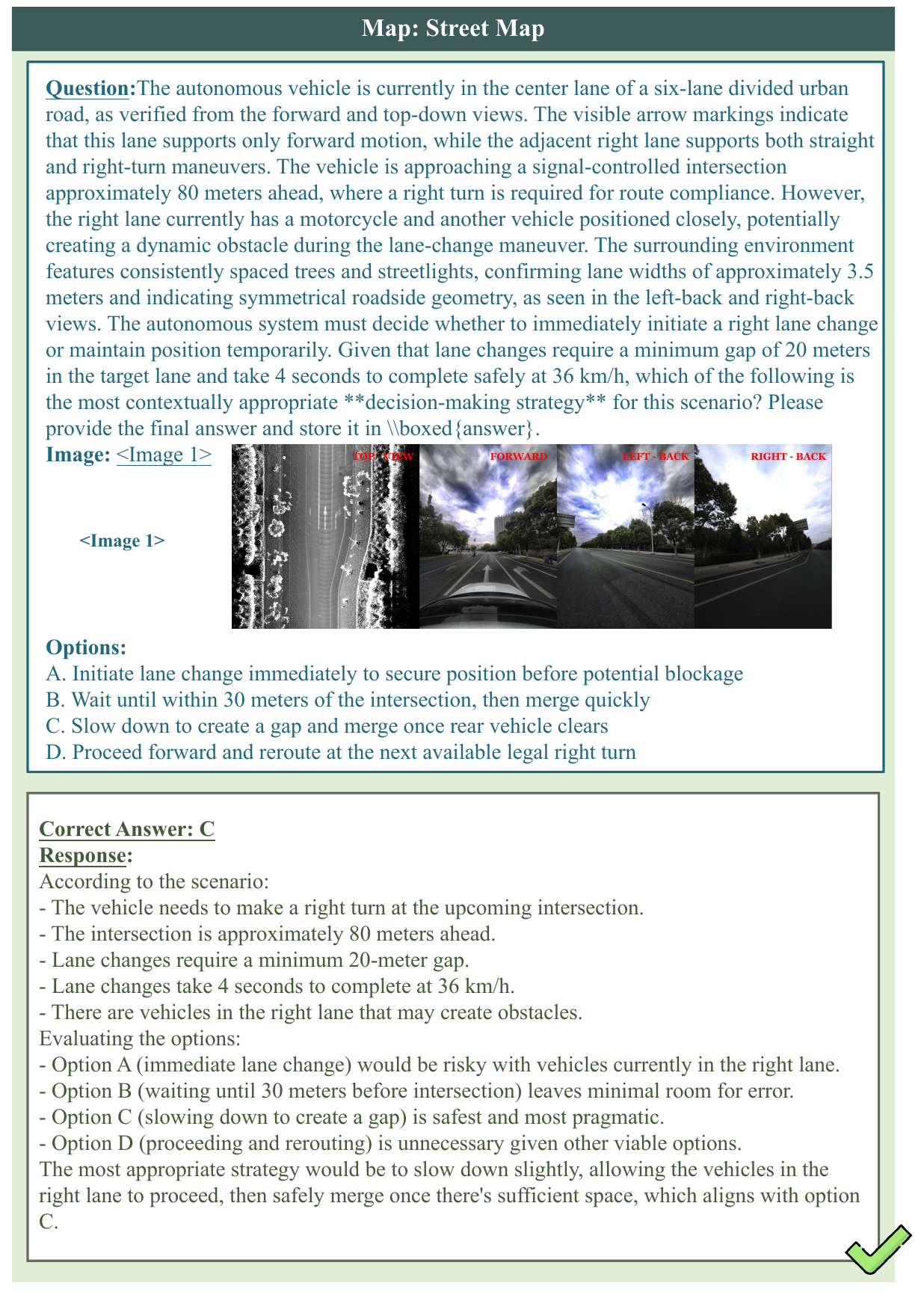}
    \vspace{-5mm}
    \caption{Street Map: Non-Thinking Case}
    \label{fig:Street Map}
\end{figure*}

\newpage
\begin{figure*}[hp]
    \centering
    \includegraphics[width=1\linewidth]{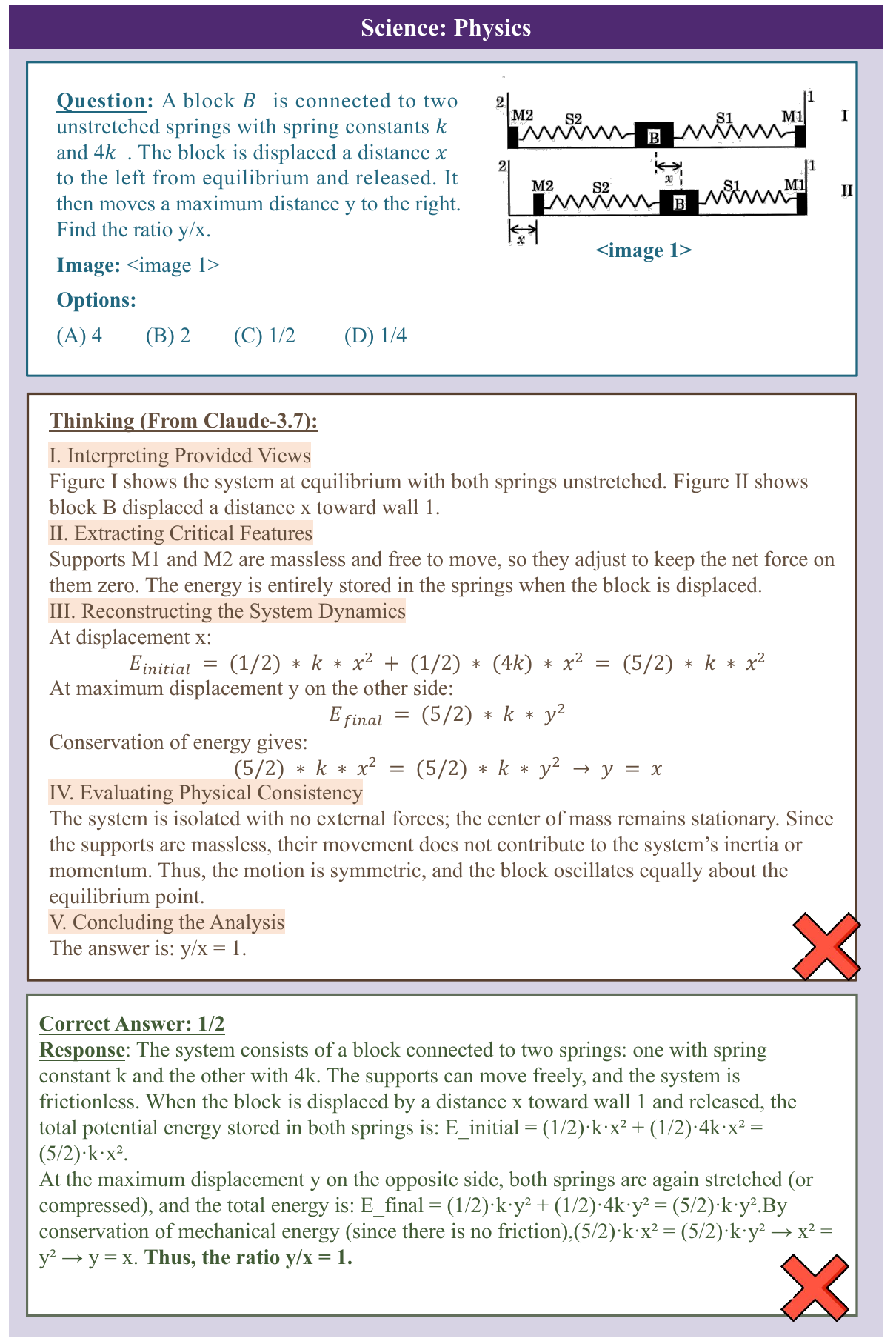}
    \vspace{-5mm}
    \caption{Science: Thinking Case}
    \label{fig:Science-Thinking}
\end{figure*}

\newpage
\begin{figure*}[hp]
    \centering
    \includegraphics[width=1\linewidth]{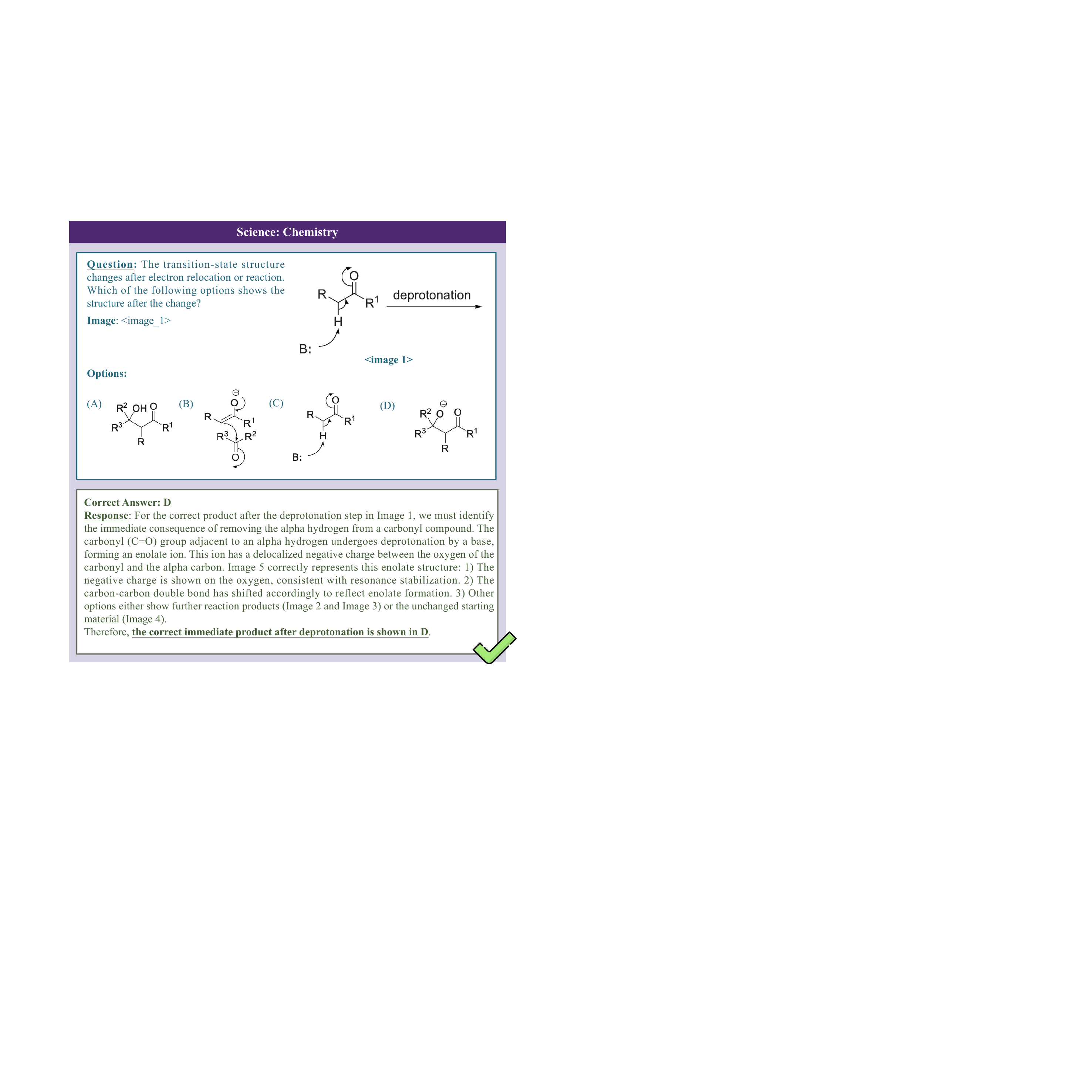}
    \vspace{-5mm}
    \caption{Chemistry: Non-Thinking Case}
    \label{fig:Chemistry}
\end{figure*}

\newpage
\begin{figure*}[hp]
    \centering
    \includegraphics[width=1\linewidth]{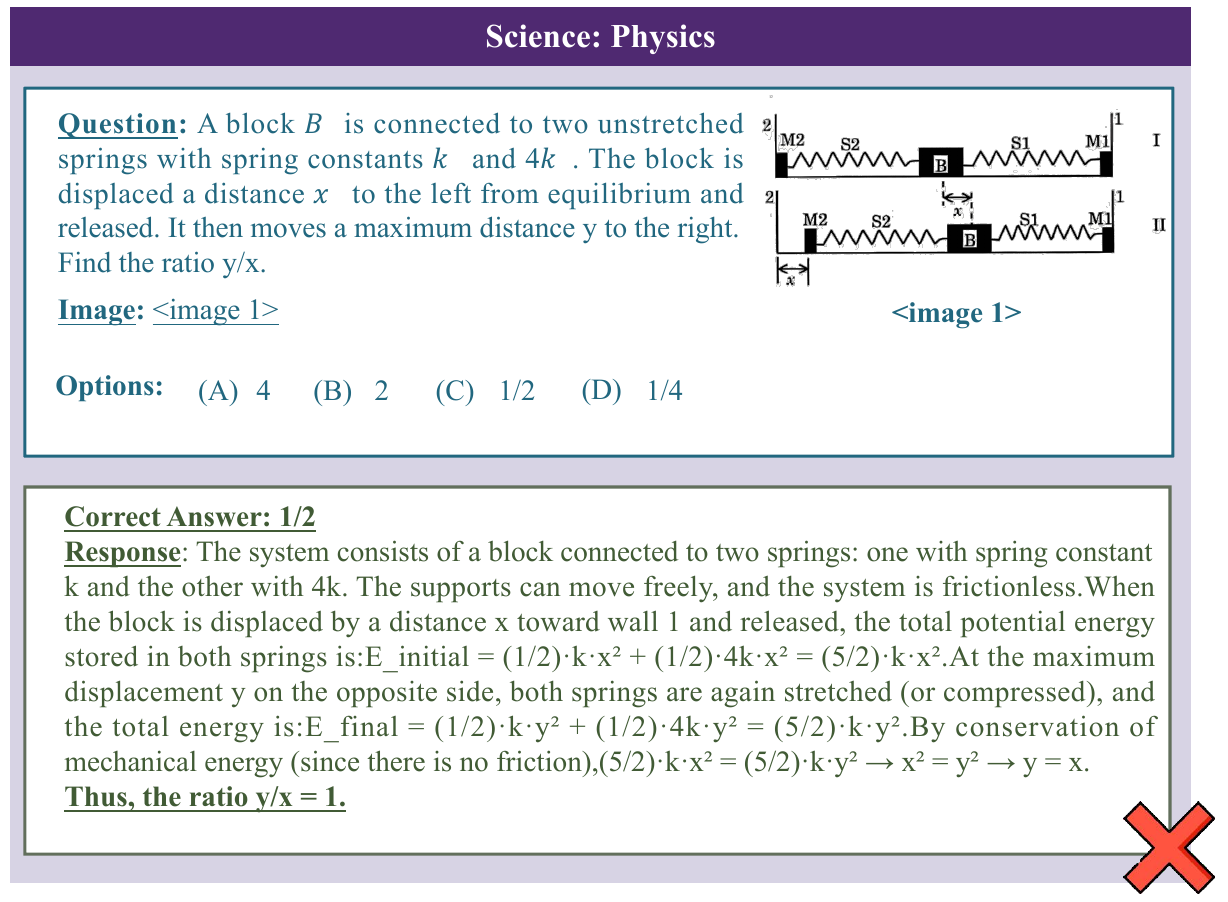}
    \vspace{-5mm}
    \caption{Physics: Non-Thinking Case}
    \label{fig:Physics}
\end{figure*}

\newpage
\begin{figure*}[hp]
    \centering
    \includegraphics[width=1\linewidth]{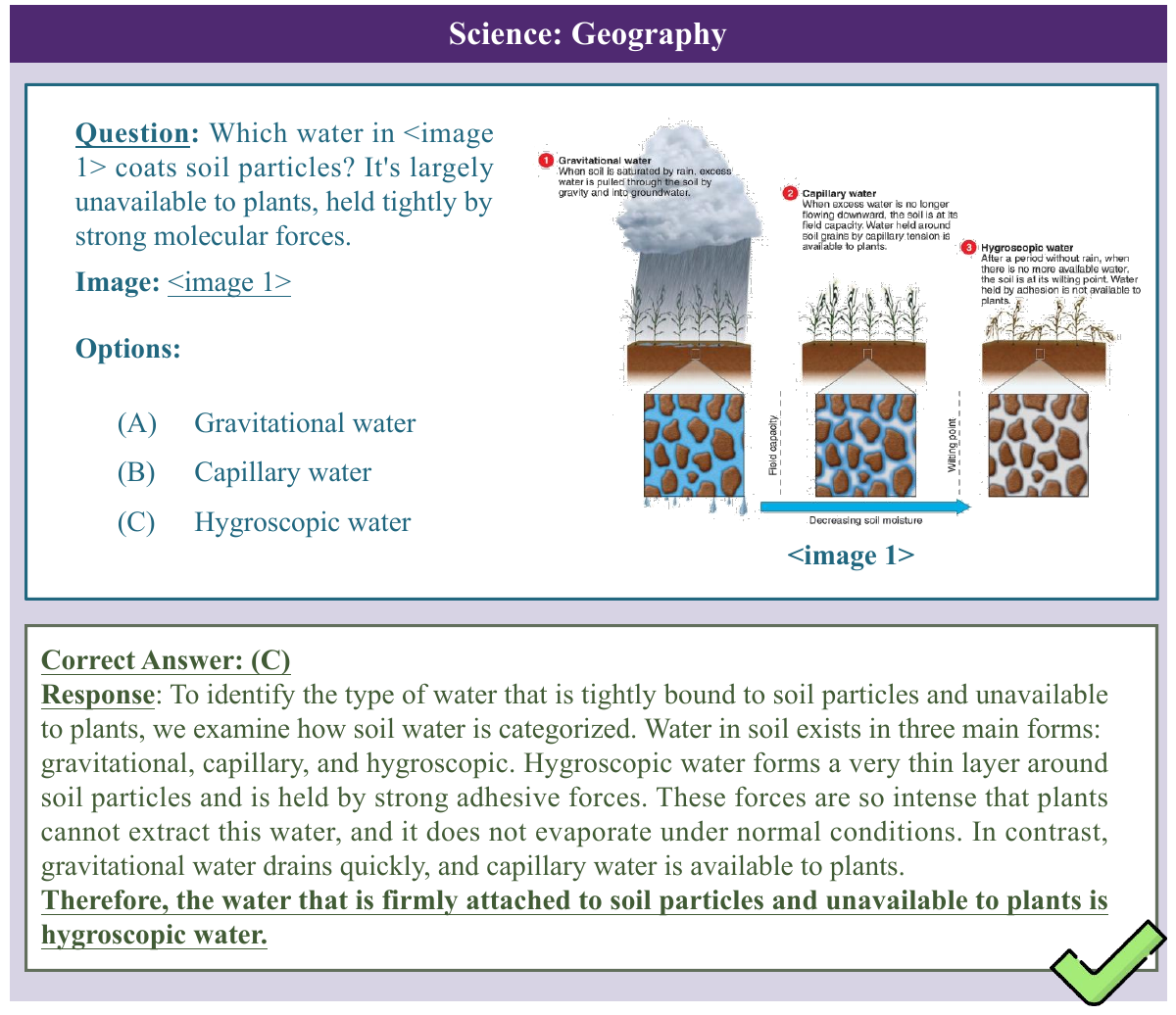}
    \vspace{-5mm}
    \caption{Geography: Non-Thinking Case}
    \label{fig:Geography}
\end{figure*}

\newpage
\begin{figure*}[hp]
    \centering
    \includegraphics[width=1\linewidth]{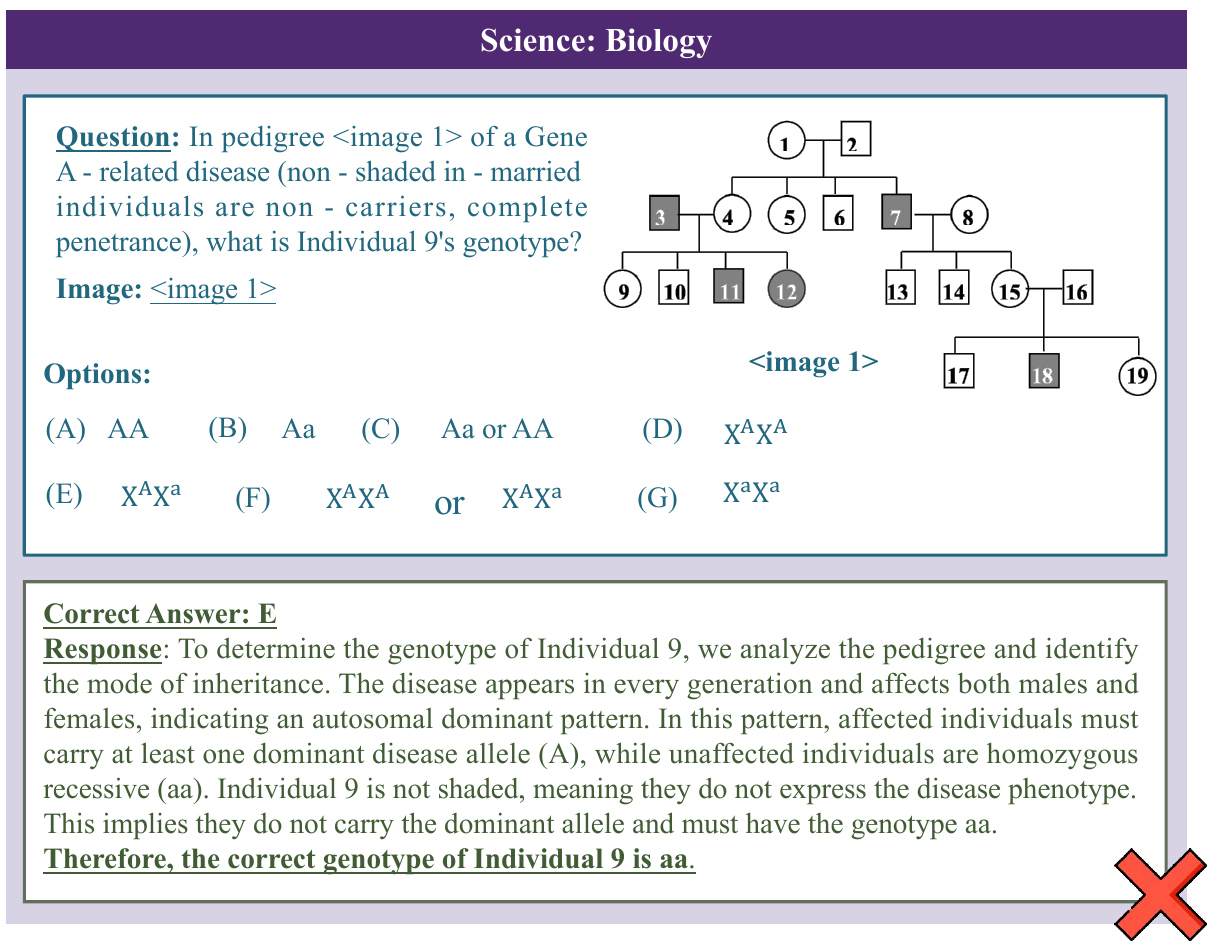}
    \vspace{-5mm}
    \caption{Biology: Non-Thinking Case}
    \label{fig:Biology}
\end{figure*}

\end{document}